\documentclass{IEEEtaes}
\usepackage{amsmath,amsfonts,amssymb}
\usepackage{array}
\usepackage{booktabs}
\usepackage{cite}
\usepackage{graphicx}
\usepackage{multirow}
\usepackage{needspace}
\usepackage{pifont}
\usepackage{stfloats}
\usepackage{textcomp}
\usepackage{url}
\usepackage{hyperref}
\usepackage[capitalize,noabbrev]{cleveref}

\jvol{XX}
\jnum{XX}
\jmonth{XXXXX}
\paper{0000000}
\pubyear{2026}
\doiinfo{TAES.2026.Doi Number}
\crefname{figure}{Fig.}{Figs.}
\crefname{table}{Table}{Tables}
\crefname{section}{Section}{Sections}
\hypersetup{
	pdftitle={P-SER: Physical Consistency-Constrained Scattering Expandable Representation for Class-Incremental HRRP Target Recognition},
	pdfauthor={Fangyi Wang and Haipeng Wang}
}
\begin{document}
	
\title{P-SER: Physical Consistency-Constrained Scattering Expandable Representation for Class-Incremental HRRP Target Recognition}
	
\author{Fangyi Wang}
\member{Graduate Student Member, IEEE}
\affil{Fudan University, Shanghai 200433, China}

\author{Peng Zhao}
\member{Graduate Student Member, IEEE}
\affil{Fudan University, Shanghai 200433, China}

\author{Haipeng Wang}
\member{Senior Member, IEEE}
\affil{Fudan University, Shanghai 200433, China}

\receiveddate{Manuscript received XXXXX 00, 0000; revised XXXXX 00, 0000; accepted XXXXX 00, 0000.\\
This work was supported in part by the National Natural Science Foundation of China (Grant No. 62271153) and the Fundamental and Interdisciplinary Disciplines Breakthrough Plan of the Ministry of Education of China (Grant No. JYB2025XDXM110).}

\corresp{\textit{(Corresponding author: Haipeng Wang.)}}

\authoraddress{The authors are with the Key Laboratory of Information Science of Electromagnetic Waves (MOE), and the Discipline and Technology Center of Microwave Vision Intelligent Sensing, Fudan University, Shanghai 200433, China
(e-mail: \href{mailto:wangfy23@m.fudan.edu.cn}{wangfy23@m.fudan.edu.cn}; \href{mailto:24210720332@m.fudan.edu.cn}{24210720332@m.fudan.edu.cn}; \href{mailto:hpwang@fudan.edu.cn}{hpwang@fudan.edu.cn}).}

\markboth{WANG ET AL.}{P-SER: PHYSICAL CONSISTENCY-CONSTRAINED SCATTERING EXPANDABLE REPRESENTATION}

\maketitle

\begin{abstract}
A high-resolution range profile (HRRP) describes target scattering responses along the radar line of sight and is widely used for radar target recognition. In long-term operation, previously unseen target classes may appear after deployment, while complete historical data may no longer be available. Class-incremental HRRP recognition therefore aims to extend a recognition model to these classes while maintaining its ability to recognize previously learned classes. This task faces three practical difficulties. First, the aspect sensitivity of HRRP imaging causes substantial changes in scattering-peak locations and amplitudes, producing large intra-class variations and old-class representation drift. Second, reflector groups can generate strong responses similar to those of ships, reducing the differences between classes and increasing misclassification. Third, conventional dynamic expansion methods duplicate complete feature extractors, leading to increasing parameter and memory costs. To address these difficulties, a Physical Consistency-Constrained Scattering Expandable Representation framework, termed P-SER, is proposed. P-SER reconstructs dominant scattering points from complex-valued HRRPs acquired through the sum and azimuth-difference channels, together with the observation geometry. The resulting spatial structure, target extent, and energy distribution are used to construct class-level scattering prototypes and sample-level saliency priors, providing stable physical references for feature learning and old-class retention. Next, a primitive scatterer dictionary describes the mechanisms of local strong-scattering responses and selectively calibrates uncertain ship--reflector predictions at inference. Finally, lightweight residual adapters are added to a shared HRRP feature backbone at each incremental stage, supporting adaptation to incoming classes with limited model growth. P-SER thus incorporates stable and interpretable scattering properties into incremental knowledge construction, retention, and inference calibration. Experiments on simulated and measured HRRP datasets show that P-SER achieves competitive recognition performance without storing old-class exemplars, while maintaining physical consistency, interference robustness, knowledge retention, and model scalability.
\end{abstract}

	
\section{Introduction}
\label{sec:introduction}

A high-resolution range profile (HRRP) records the range projection of a target's scattering responses in a compact one-dimensional form and supports low-latency target recognition. HRRP recognition systems increasingly operate in open and continuously evolving environments, where new target categories may appear after deployment. Meanwhile, complete historical HRRP data are often unavailable because of storage, transmission, or operational constraints. Class-incremental HRRP recognition addresses this setting by learning newly observed classes in successive stages while retaining the ability to discriminate all previously encountered classes~\cite{Rebuffi2017,Li2017,Douillard2020,Yan2021}. This capability is important for long-term maritime surveillance and other time-sensitive applications, since repeatedly collecting all historical observations and retraining a model from scratch is generally impractical. A conventional recognizer updated only with newly arriving samples tends to overwrite previously acquired representations, resulting in catastrophic forgetting during sustained operation.

\begin{figure}[t]
	\centering
	\includegraphics[width=\columnwidth]{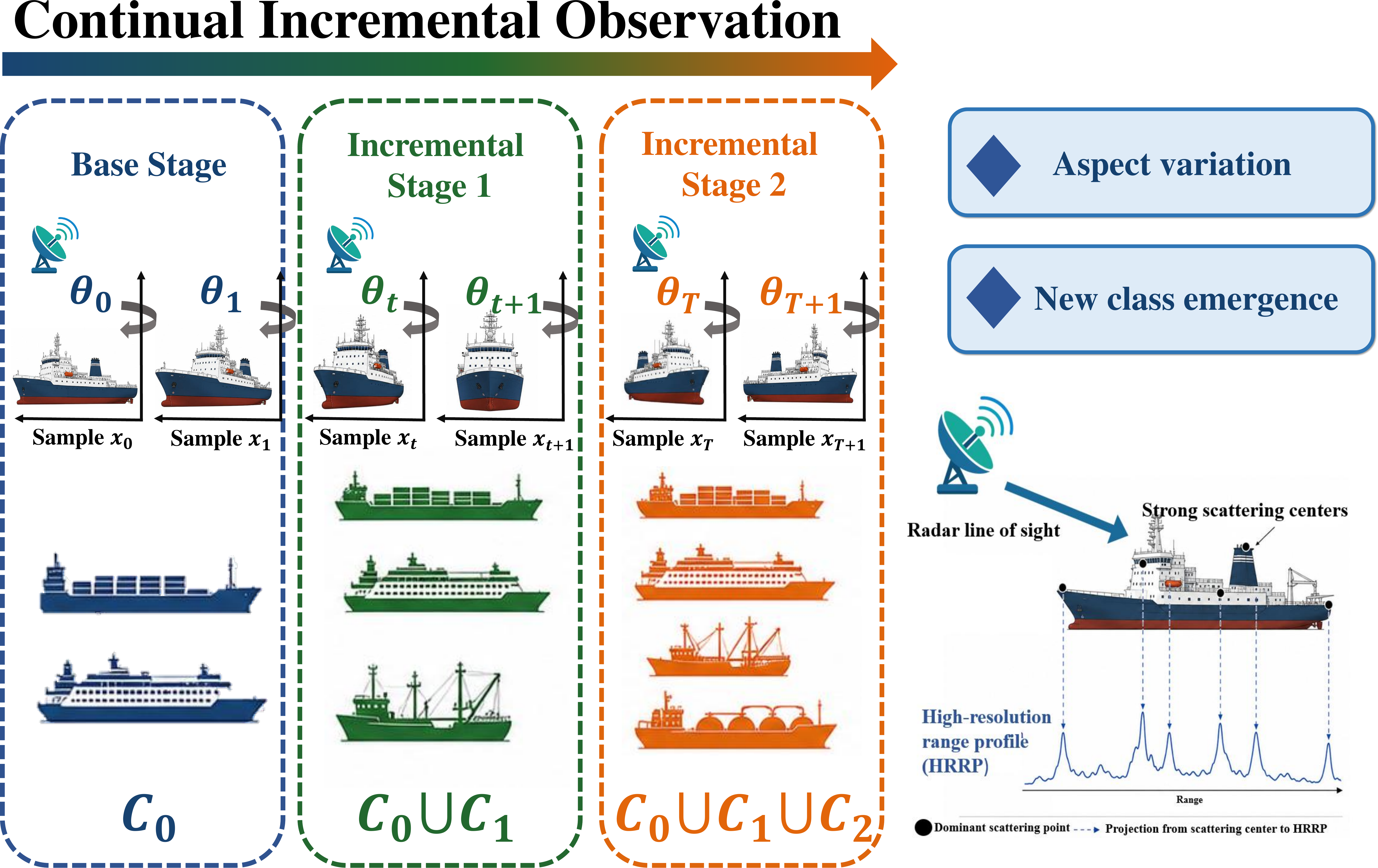}
	\caption{Illustration of continual class-incremental HRRP observation and its physical scattering response. Target classes expand progressively, while the HRRP imaging mechanism remains intrinsically sensitive to viewing aspect. Dominant responses are formed by the range projection of physical scattering centers, motivating the use of stable physical properties for incremental recognition.}
	\label{fig:hrrp_generation_incremental}
\end{figure}

Although class-space expansion is the defining change in the considered task, the HRRP data acquired at each stage are also affected by the imaging mechanism and scene interference, as illustrated in \Cref{fig:hrrp_generation_incremental}. Considering both the knowledge-retention requirement of class-incremental learning and the deployment constraints of radar systems, the main difficulties of class-incremental HRRP recognition can be summarized as representation instability, interference vulnerability, and model inefficiency.
The first challenge originates from the intrinsic aspect sensitivity of HRRP imaging. As the radar line of sight varies among observations, the range projections, mutual occlusion, and electromagnetic responses of target scattering components change accordingly. Consequently, the locations and amplitudes of dominant scattering peaks may fluctuate substantially for the same physical target, producing large intra-class variations. This property increases the difficulty of class-incremental recognition because the model must learn new classes without full access to historical samples that describe the diverse aspect-dependent appearances of old classes. A purely data-driven model may associate class identity with transient peak patterns and consequently suffer from feature drift. Such representations also provide limited insight into which intrinsic target properties have been retained.

The second challenge arises from strong-scattering interference that may appear or change across observation stages. Reflector groups, for example, can generate a small number of highly concentrated peaks whose range locations and amplitudes resemble the dominant responses of ship HRRPs. These responses may even occupy most of the profile energy. A classifier relying mainly on range-domain peak patterns may therefore confuse ships with reflector groups, although their underlying scattering mechanisms are fundamentally different. This ambiguity exposes a limitation of correlation-driven recognition: similar HRRP peak configurations do not necessarily correspond to the same physical scattering composition.

The third challenge concerns deployment efficiency. Continually operating HRRP recognition systems require timely model updates under restricted storage and computational resources. Dynamic network expansion provides useful representational plasticity for new classes, but conventional approaches often duplicate a complete feature extractor at each incremental stage. Their parameter and memory costs therefore increase rapidly as the class sequence grows. Such expansion is unsuitable for long-term radar deployment, where new-class adaptability, old-class retention, and model compactness are simultaneously required.

The first two challenges reveal a common limitation of purely data-driven HRRP recognition. Features learned from the current observation distribution do not explicitly preserve the stable physical properties of the target. To address this limitation, physical consistency is introduced into class-incremental HRRP recognition. Dominant scattering points are first reconstructed from the complex sum- and difference-channel profiles together with the observation geometry. Their spatial structure, target extent, and energy distribution are more stable across viewing aspects than directly observed HRRP peak patterns. These physical attributes are subsequently constructed as class-level scattering prototypes and sample-level saliency priors. They anchor feature learning to the target itself rather than to a particular aspect-dependent observation, thereby reducing representation drift and supporting old-class retention.

Physical scattering mechanisms are also incorporated at inference. A primitive scatterer dictionary (PSD) characterizes the mechanisms associated with local strong-scattering components. When the HRRP recognition network is uncertain between a ship and a reflector group, the PSD mechanism evidence extracted from dominant-peak neighborhoods is used to calibrate the original prediction selectively. The final decision therefore depends on both the learned HRRP representation and the physical origin of the local response, improving robustness against strong-scattering interference.

For efficient expansion, a shared HRRP feature backbone is retained throughout the class sequence, and only a compact residual adapter is introduced at each incremental stage. The adapters provide stage-specific plasticity for learning new-class characteristics without repeatedly duplicating the complete feature extraction network. This design limits model growth while preserving previously acquired knowledge.

\begin{figure}[t]
	\centering
	\includegraphics[width=\columnwidth]{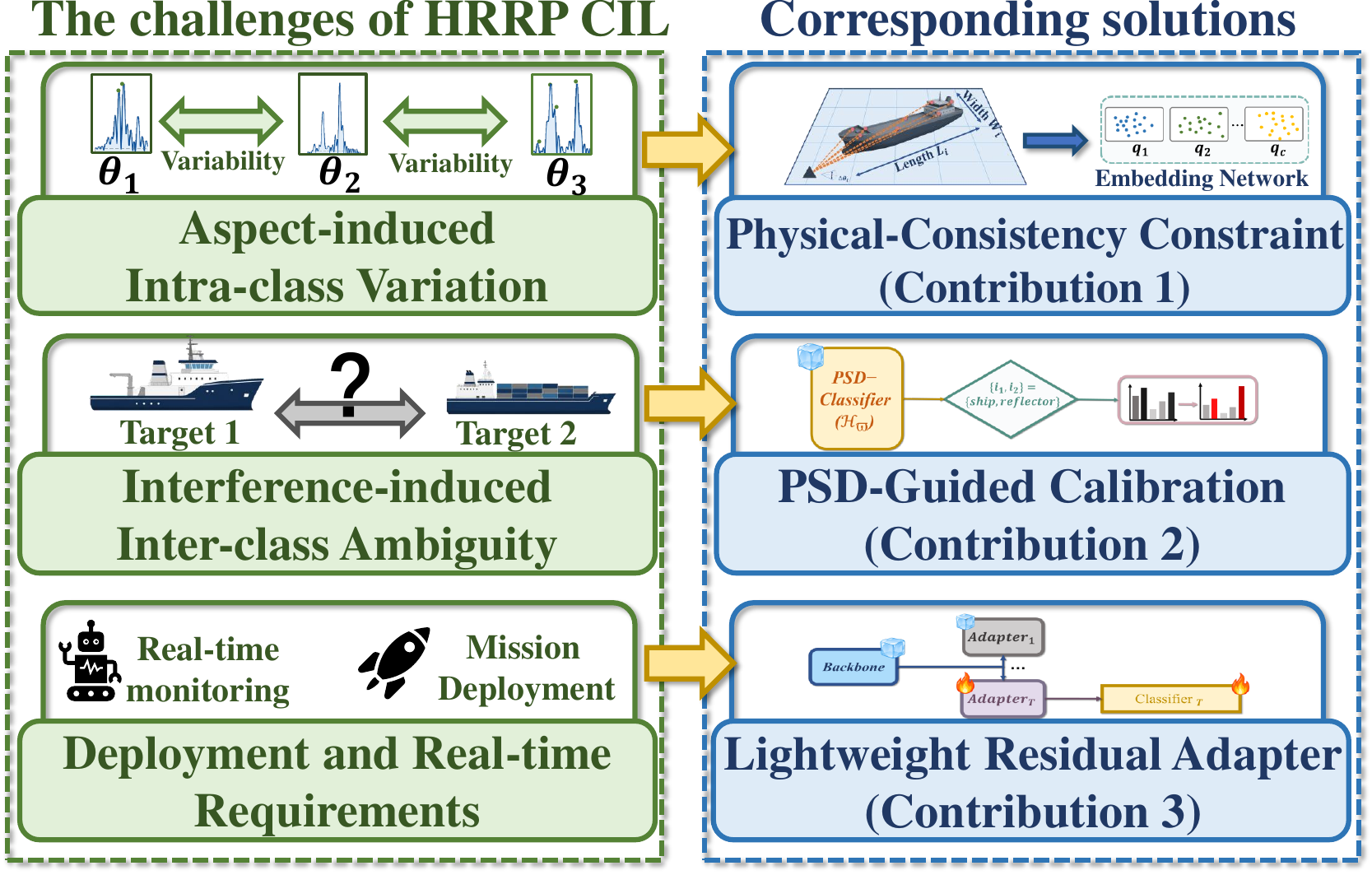}
	\caption{Key challenges in class-incremental HRRP recognition and the corresponding solutions in P-SER.}
	\label{fig:12}
\end{figure}

Based on these considerations, a Physical Consistency-Constrained Scattering Expandable Representation framework, termed P-SER, is proposed for class-incremental HRRP recognition, as illustrated in \Cref{fig:12}. Unlike methods that rely solely on statistical correlations in HRRP observations, P-SER allows stable and interpretable target scattering properties to participate directly in incremental knowledge construction, retention, and inference calibration. The main contributions are summarized as follows:

\begin{enumerate}
	\item A physical-consistency scattering prototype constraint is proposed to alleviate representation drift caused by the intrinsic aspect sensitivity of HRRP observations. Dominant scattering points and target attributes are reconstructed from the complex sum- and difference-channel profiles and observation geometry. They are incorporated into class-level prototypes and sample-level saliency priors, providing stable physical references for feature learning and old-class retention across viewing aspects.
	
	\item A PSD-guided inference calibration method is developed for strong-scattering interference. It analyzes the physical scattering mechanisms associated with dominant local HRRP responses and selectively corrects ambiguous ship--reflector predictions without modifying the training procedure of the original recognition network.
	
	\item Lightweight residual adapters are introduced for parameter-efficient incremental expansion. Compact stage-specific branches learn new-class HRRP representations on a shared backbone, limiting parameter growth while maintaining previously acquired knowledge.
\end{enumerate}

Experiments on simulated and measured HRRP datasets demonstrate the effectiveness and efficiency of P-SER. The proposed method achieves state-of-the-art average recognition performance on both datasets, exceeding the baseline by 26.55\% and 19.05\%, respectively. Compared with the second-best methods, P-SER further improves the average accuracy by 2.14\% and 1.61\% while using only 19.2\% and 33.4\% of their parameters. These results demonstrate that P-SER achieves a favorable balance among recognition accuracy, resistance to catastrophic forgetting, robustness against strong-scattering interference, and model complexity.
\section{Related Work}
\subsection{HRRP Target Recognition}
Before the advent of deep learning, HRRP target recognition primarily relied on statistical models and handcrafted descriptors, including radial extent, peak count, power spectra, and multiaspect dependencies~\cite{Pilcher2011,Du2015,Du2008,Zhu2007,Kim2002,Slomka1999}. These methods are computationally efficient and physically interpretable, but their performance depends strongly on signal quality and expert feature design. With the development of deep learning, hierarchical representations extracted directly from range profiles have increasingly replaced or integrated these handcrafted characteristics.

In recent years, convolutional, recurrent, and generative models have substantially advanced radar target recognition by learning latent characteristics from such data~\cite{Liao2018,Wan2020,Guo2019,Li2025e,Han2026}. However, the representations obtained by these methods mainly rely on statistical correlations in the observed data. Under aspect-dependent response variation, these correlations may be unstable and offer limited physical interpretability.

Convolutional neural network (CNN)-based methods exploit local scattering patterns and their multiscale combinations. Liao \textit{et al.}~\cite{Liao2018} proposed a concatenated deep architecture for radar HRRP recognition. Wan \textit{et al.}~\cite{Wan2019} combined a convolutional encoder with a deconvolutional decoder to integrate target recognition and outlier rejection, while Guo \textit{et al.}~\cite{Guo2019} developed a deep one-dimensional residual-inception network. More recent studies have incorporated radar-domain knowledge into feature learning. Zhou \textit{et al.}~\cite{Zhou2024} introduced knowledge-guided bidirectional temporal modeling and prototype enhancement, and Liu \textit{et al.}~\cite{Liu2025} proposed a target-aspect-guided network with geometric constraints for imbalanced HRRP recognition.

Recurrent neural network (RNN)-based methods use the sequential organization of HRRP range cells to model long-range dependencies~\cite{Zhang2020,Zeng2022,Su2023}. Chen \textit{et al.}~\cite{Chen2021} combined tensor recurrent modeling with a Bayesian nonparametric mixture to capture complex nonstationary distributions. Guo \textit{et al.}~\cite{Guo2020a} proposed a variational temporal deep generative model, whereas Chen \textit{et al.}~\cite{Chen2021a} developed a bidirectional recurrent gamma belief network that models correlations in both range directions.

Generative HRRP recognition methods mainly employ variational autoencoders (VAEs) and generative adversarial networks (GANs). In a VAE, the encoder learns a latent distribution and the decoder reconstructs the observation~\cite{Kingma2013}. Du \textit{et al.}~\cite{Du2019} proposed a factorized discriminative conditional VAE to reduce the mismatch between feature learning and classification. Liao \textit{et al.}~\cite{Liao2021} factorized the decoder according to target classes, and Chen \textit{et al.}~\cite{Chen2020a} introduced a mixture structure for more flexible data modeling. Recognition-aware adversarial generation~\cite{Huang2021}, small-sample generation under noise~\cite{Yin2024}, and continuous learning based on a CVAE--GAN~\cite{Li2023} further demonstrate the value of generative representations. Recent studies have also explored synthetic-to-real HRRP construction and structure-aware generative modeling for high-fidelity profile synthesis~\cite{Li2026,Li2026a}. Nevertheless, these methods are primarily designed for closed-set recognition and do not directly address the retention of historical classes under sequential class arrival.

Beyond conventional closed-set classification, recent HRRP research has increasingly considered open-set recognition and unknown-target rejection through statistical tail modeling, prototype-based representation, and sequence-aware decision mechanisms~\cite{Xia2023,Chen2025,Li2025d,Zhang2025,Wang2025}. Meanwhile, robust HRRP recognition under domain shifts, corrupted observations, cross-frequency variations, adversarial disturbances, and active interference has also received growing attention~\cite{Yang2025,Zhang2023a,Zhang2026,Liu2025a,Yang2026,Wang2026,Sun2026}. These developments extend radar recognition to increasingly complex operating conditions, but physical consistency, interference discrimination, and parameter-efficient class-incremental adaptation have rarely been considered within a unified framework.

\subsection{Class-Incremental Learning}
Class-incremental learning handles a sequence of stages in which disjoint groups of classes are introduced over time, while a single classifier is evaluated over all classes observed so far~\cite{Rebuffi2017}. Its central challenge is to learn discriminative representations for new categories without overwriting previously acquired knowledge. Existing methods can be broadly grouped into rehearsal, regularization or distillation, dynamic expansion, and exemplar-free learning.

Rehearsal-based methods preserve a bounded set of historical samples and replay them during subsequent updates. iCaRL~\cite{Rebuffi2017} established a representative framework by combining exemplar rehearsal, representation learning, and prototype-based classification. Later studies improved exemplar selection through uncertainty or data-stream characteristics~\cite{Bang2022}, while De Lange \textit{et al.}~\cite{DeLange2021} investigated continual prototype evolution under nonstationary streams. Rehearsal is effective because it directly approximates the historical distribution, but its performance depends on the permission and memory available for retaining old samples.

Knowledge distillation (KD)~\cite{Hinton2015} and parameter regularization provide another route for transferring historical knowledge. Learning without Forgetting~\cite{Li2017} constrained the outputs of the updated model using responses from its previous version. Kirkpatrick \textit{et al.}~\cite{Kirkpatrick2017} used Fisher information to penalize changes to important parameters, while remote-sensing-oriented methods further constrained representation drift~\cite{Tai2023,Chen2020}. PODNet~\cite{Douillard2020} preserves spatially pooled intermediate representations, and Kang \textit{et al.}~\cite{Kang2022} proposed importance-aware feature consolidation. These approaches avoid unbounded expansion but may still struggle when new classes cause large representation shifts.

Dynamic expansion allocates additional capacity to incoming classes and has achieved strong performance in challenging CIL settings. Dynamically expandable networks~\cite{Lee2017} expand and retrain task-relevant neurons while using sparsity constraints to control redundancy. DER~\cite{Yan2021} further introduced task-specific representation branches, demonstrating the benefit of isolating new-class feature learning. FOSTER~\cite{Wang2022} combines feature boosting with compression to regulate the expanded model. TagFex~\cite{Zheng2025} learns task-agnostic features through an auxiliary model and transfers them to task-specific branches to reduce feature collision. Although expansion generally improves plasticity, repeatedly introducing complete feature extractors incurs substantial storage cost.

When historical data cannot be retained, exemplar-free methods reconstruct or approximate old-class representations without storing raw samples. FeTrIL~\cite{Petit2023} combines a fixed feature extractor with feature translation to synthesize pseudo-features for old classes. DS-AL~\cite{Zhuang2024} uses an analytic linear solution together with a compensation stream to mitigate underfitting. Such methods provide an attractive balance between privacy, storage, and recognition accuracy, but their effectiveness depends strongly on the stability of the feature space inherited from the base stage.

\subsection{HRRP Incremental Learning}
As discussed in \Cref{sec:introduction}, directly transferring image-oriented CIL algorithms to HRRP recognition does not fully exploit the one-dimensional scattering structure of the observations. Only a limited number of studies have specifically considered HRRP incremental recognition. Li \textit{et al.}~\cite{Li2025a} studied an exemplar-free setting and combined a local scattering patch multilayer perceptron with Gaussian-truncated Mahalanobis classification. Li \textit{et al.}~\cite{Li2025c} used pooled distillation to transfer knowledge into a lightweight feature extractor. Chang \textit{et al.}~\cite{Chang2025} combined a task-agnostic autoencoding branch, expandable task-specific branches, and multilevel KD to control model growth under storage constraints.

These studies incorporate range-profile structure and exemplar-free requirements into incremental recognition, but the robustness of the learned representation to pronounced aspect changes and strong-scattering interference remains insufficiently explored. In practical applications, abrupt aspect variation can cause severe migration and amplitude fluctuation of dominant peaks, while reflector groups can produce responses similar to those of ship structures. Moreover, full task-specific expansion remains costly as the number of stages increases. These limitations motivate an HRRP-oriented CIL framework that jointly exploits stable physical properties, scattering-mechanism information, and lightweight expandable representations.
	
\section{Methodology}
\begin{figure*}[t]
	\centering
	\includegraphics[width=\textwidth]{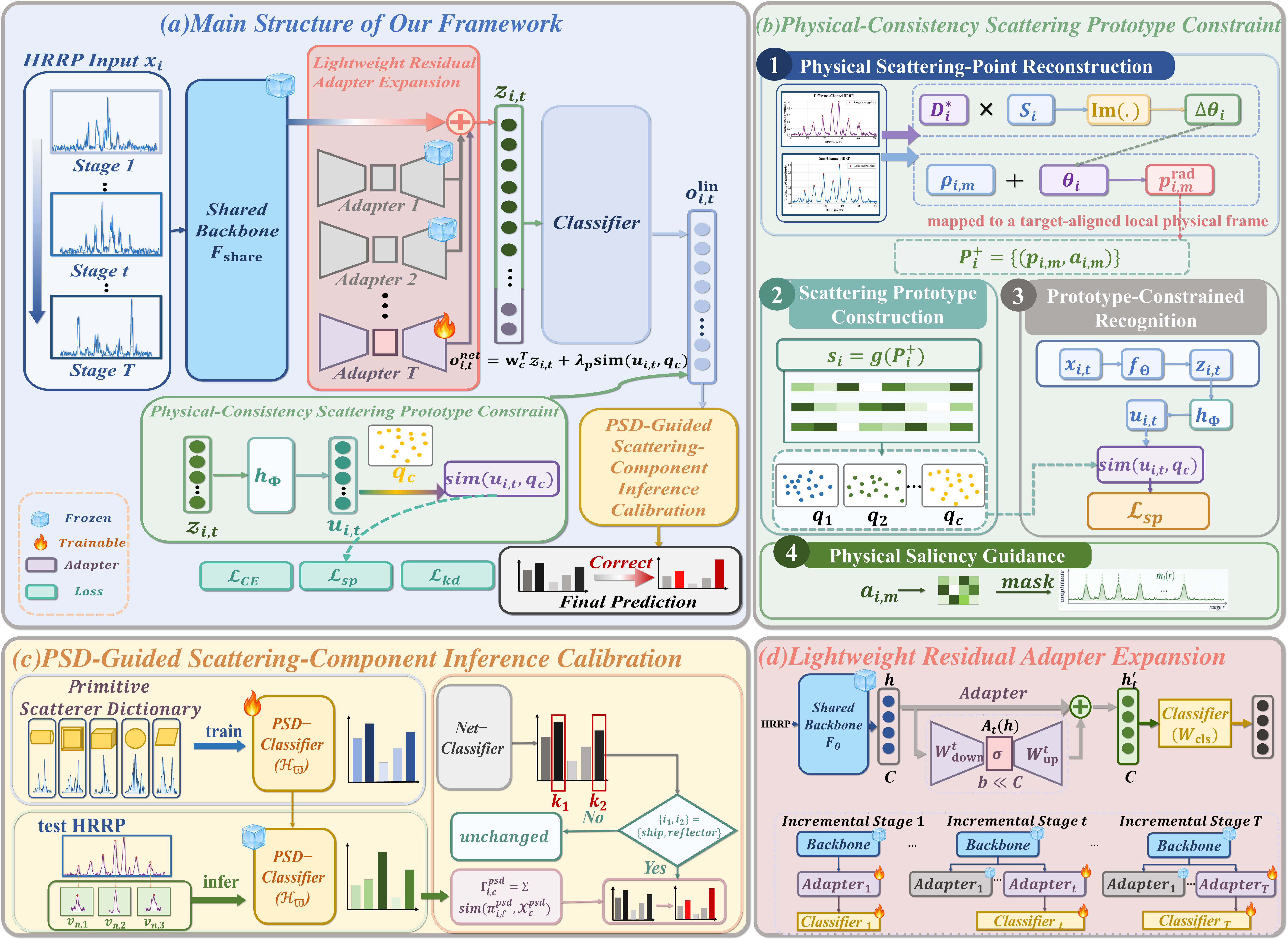}
	\caption{Overall framework of the proposed P-SER. The framework consists of three complementary components: physical-consistency scattering prototype constraint (PCSP), PSD-guided scattering-component inference calibration (PSD-SCIC), and lightweight residual adapter expansion (LRAE).}
	\label{fig:main}
\end{figure*}
The base stage is indexed by $t=0$, and subsequent incremental updates by $t\geq1$. At stage $t$, only the current training set
\begin{equation}
	\mathcal{D}^{t}=\{(\mathbf{x}_i,y_i)\}_{i=1}^{N_t},
	\qquad y_i\in\mathcal{C}^{t},
\end{equation}
is available, where $\mathbf{x}_i=[\mathbf{S}_i;\mathbf{D}_i]\in\mathbb{C}^{2\times N_r}$ contains the complex sum- and azimuth-difference-channel profiles with $N_r$ range cells, $y_i$ is its label, $N_t$ is the number of current-stage samples, and $\mathcal{C}^{t}$ is the class set introduced at stage $t$. Class sets are disjoint across stages, and no old-class exemplars are retained. After stage $t$, the model must classify all classes in $\mathcal{C}^{0:t}=\bigcup_{j=0}^{t}\mathcal{C}^{j}$.

As illustrated in \Cref{fig:main}, P-SER addresses this problem through three complementary components. PCSP reconstructs dominant scattering points and converts their physical attributes into class-level prototypes and sample-level saliency priors. PSD-SCIC provides an inference-time mechanism check for ambiguous ship--reflector predictions. LRAE expands representation capacity through compact residual adapters rather than complete backbone copies.

\subsection{Physical-Consistency Scattering Prototype Constraint}
\label{subsec:pcsp}
\begin{figure}[t]
	\centering
	\includegraphics[width=\columnwidth]{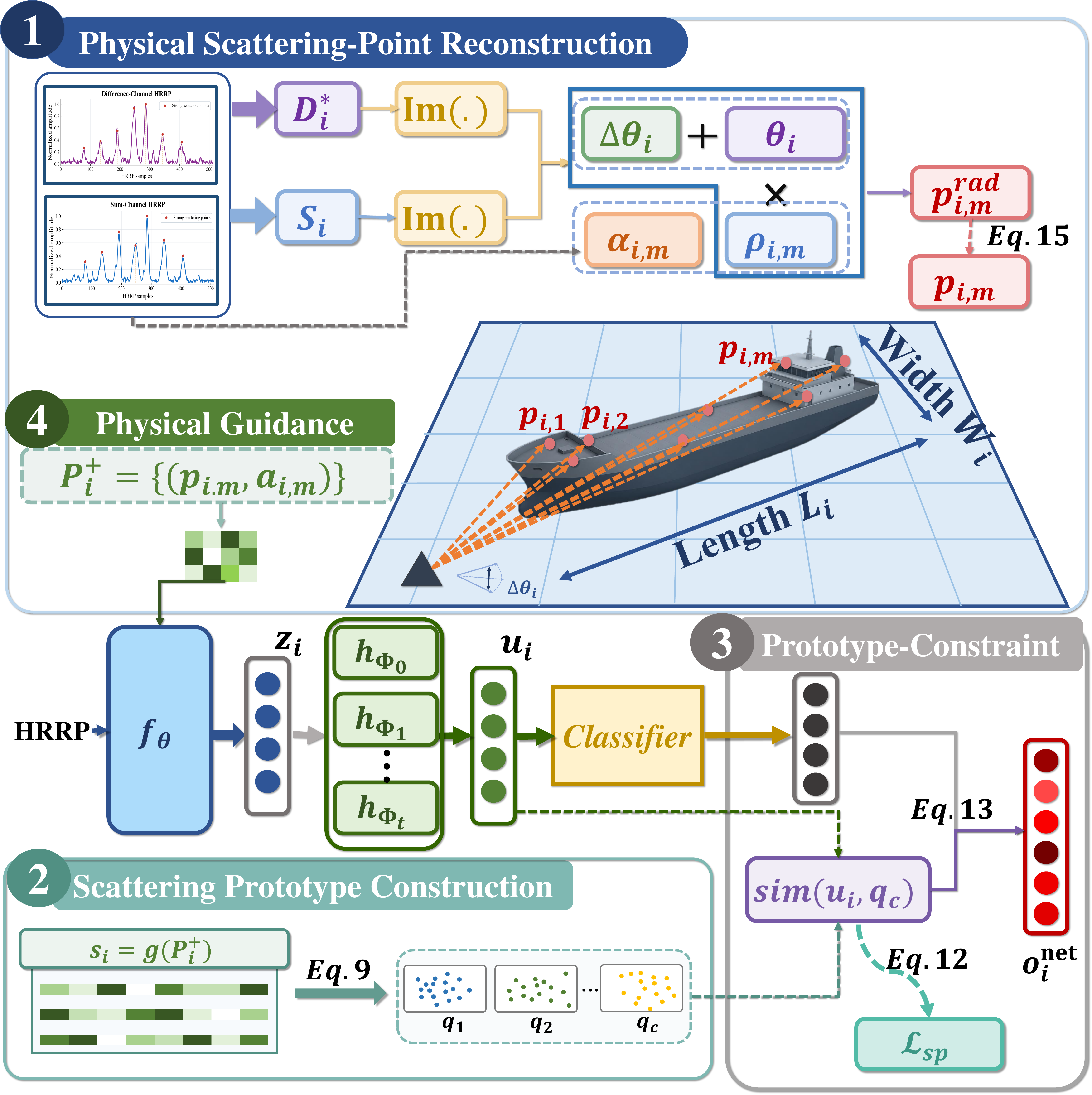}
	\caption{Overview of PCSP, which reconstructs physically consistent scattering points from HRRP observations and uses the resulting class-level prototypes and sample-level saliency priors to constrain incremental recognition.}
	\label{fig:31}
\end{figure}
As illustrated in \Cref{fig:31}, dominant HRRP peaks are produced by target-dependent scattering centers rather than arbitrary range-domain fluctuations, although their observed locations vary with the viewing geometry. PCSP establishes their physical correspondence through calibrated monopulse angle estimation, radar-coordinate reconstruction, and data-derived spatial alignment. The resulting sparse point set is represented in a centered local frame whose principal axis is estimated from the reconstructed points; neither the true target heading nor class-specific metadata is required. Its stable geometric and energy attributes are subsequently used as physical anchors during incremental learning.

\subsubsection{Physical Scattering-Point Reconstruction}

For the $i$-th observation, let $S_i^{\mathrm{raw}}(r)$ and $D_i^{\mathrm{raw}}(r)$ denote the raw complex sum-channel and azimuth-difference-channel responses at range cell $r$, respectively. The two channels are first compensated using the complex calibration responses $G_{\Sigma}(r)$ and $G_{\Delta}(r)$ of the corresponding radar configuration:
\begin{equation}
	S_i(r)=\frac{S_i^{\mathrm{raw}}(r)}{G_{\Sigma}(r)},
	\qquad
	D_i(r)=\frac{D_i^{\mathrm{raw}}(r)}{G_{\Delta}(r)},
	\label{eq:channel_calibration}
\end{equation}
Here, $G_{\Sigma}(r)$ and $G_{\Delta}(r)$ are obtained from a reference-source calibration measurement and compensate the relative amplitude and phase responses of the sum and difference receiving paths. After this compensation, the quadrature convention adopted in this work gives the main-lobe monopulse relation $D_i(r)/S_i(r)\simeq-\mathrm{j}\kappa_{\mathrm{mp}}\Delta\theta_i(r)$~\cite{Sherman2011,Nickel2006}, where $\mathrm{j}=\sqrt{-1}$, $\theta_{\mathrm{3dB}}$ is the azimuthal 3-dB beamwidth, and $\kappa_{\mathrm{mp}}$ is the monopulse slope. Under this convention, $D_i^{*}(r)S_i(r)$ has an imaginary component proportional to $\Delta\theta_i(r)|S_i(r)|^2$. The angular deviation and effective observation angle are therefore obtained jointly as
\begin{equation}
	\begin{aligned}
		\Delta\theta_i(r)
		&=
		\frac{\operatorname{Im}\{D_i^{*}(r)S_i(r)\}}
		{\kappa_{\mathrm{mp}}|S_i(r)|^2+\epsilon_{\theta}},\\
		\theta_i(r)&=\theta_i^{0}+\Delta\theta_i(r),
		\qquad
		|\Delta\theta_i(r)|\leq\frac{\theta_{\mathrm{3dB}}}{2}.
	\end{aligned}
	\label{eq:angle_error}
\end{equation}
where $(\cdot)^{*}$ denotes complex conjugation, $\operatorname{Im}\{\cdot\}$ extracts the imaginary part, and $\epsilon_{\theta}$ is a small positive constant that prevents unstable division in weak-response cells. The coefficient $\kappa_{\mathrm{mp}}$ is estimated once for each radar configuration by least-squares fitting the calibrated monopulse ratio to known angular offsets of a reference target within the main lobe. It is then fixed for all samples acquired with that configuration and does not depend on target pose or signal-to-noise ratio; a range- or frequency-dependent calibration table can be used when required by the front-end response. A different channel phase convention may place the monopulse slope on the real axis, in which case the complex calibration rotates the measured ratio to the adopted quadrature convention. Only estimates within the calibrated main-lobe interval are retained. The beam-pointing azimuth $\theta_i^{0}$ is sensor-geometry metadata available at both acquisition and recognition, rather than the target heading or a class annotation.

When several unresolved scattering centers occupy one range cell, \eqref{eq:angle_error} yields the coherent apparent angle of their composite response, which is generally dominated by the strongest contributors rather than the angle of an individually resolved center. Accordingly, each retained HRRP peak is interpreted as a dominant scattering component at the available range resolution.

The dominant scattering responses are detected from the HRRP amplitude sequence $a_i(r)=|S_i(r)|$. The candidate peak set is defined as
\begin{equation}
	\begin{aligned}
		\mathcal{M}_i=
		\bigl\{r_m\in\Omega_i\,\bigm|\;&
		a_i(r_m)=\max_{r\in\mathcal{N}(r_m)}a_i(r),\\
		& a_i(r_m)\geq
		\beta\max_{r\in\Omega_i}a_i(r)\bigr\},
	\end{aligned}
	\label{eq:peak_set}
\end{equation}
where $\Omega_i$ denotes the valid target range gate, $\mathcal{N}(r_m)$ is the local range neighborhood centered at $r_m$, $\beta$ is the relative amplitude threshold with respect to the strongest response, and $r_m$ is the range index of the $m$-th detected peak. The relative threshold suppresses weak local fluctuations before physical reconstruction while preserving dominant scattering responses. Let $r_i^{\mathrm{st}}$ be the first cell of $\Omega_i$. The slant range and corresponding radar-frame point of each retained peak are
\begin{equation}
	\begin{aligned}
		\rho_{i,m}&=R_i^{0}+\eta(r_m-r_i^{\mathrm{st}}),\\
		\mathbf{p}_{i,m}^{\mathrm{rad}}
		&=\rho_{i,m}
		[\cos\theta_i(r_m),\sin\theta_i(r_m)]^{T},
	\end{aligned}
	\label{eq:radar_point_mapping}
\end{equation}
where $R_i^{0}$ is the calibrated slant range associated with $r_i^{\mathrm{st}}$ in the radar range-gate record and $\eta$ is the range-cell spacing. With $a_{i,m}=a_i(r_m)$, the radar-frame sparse point set is $\mathcal{P}_i^{\mathrm{rad}}=\{(\mathbf{p}_{i,m}^{\mathrm{rad}},a_{i,m})\}_{m=1}^{M_i}$, where $M_i=|\mathcal{M}_i|$ is the number of candidate peaks and $a_{i,m}$ is the amplitude of the $m$-th scattering response.

Since not all local maxima correspond to stable scattering centers, a main-cluster filtering operation is applied to suppress isolated false peaks:
\begin{equation}
	\mathcal{P}_i^{\mathrm{rad},+}
	=
	\operatorname{MainCluster}(\mathcal{P}_i^{\mathrm{rad}}).
	\label{eq:main_cluster}
\end{equation}
Here, $\operatorname{MainCluster}(\cdot)$ denotes a neighborhood-consistency filtering operation that retains the dominant scattering structure supported by adjacent peaks and removes isolated responses caused by noise, sidelobe leakage, or unstable weak scattering. Let $\mathcal{K}_i$ index the retained points. Their amplitude-weighted center $\mathbf{c}_i$ and covariance $\mathbf{C}_i$ are computed from $\mathcal{P}_i^{\mathrm{rad},+}$. Denote the principal eigenvector of $\mathbf{C}_i$ by $\mathbf{v}_{i,1}$ and its orientation by $\widehat{\psi}_i=\operatorname{atan2}(v_{i,1}^{(2)},v_{i,1}^{(1)})$. The sign of $\mathbf{v}_{i,1}$ is selected to have a nonnegative projection on the radar line of sight, resolving the $180^{\circ}$ eigenvector ambiguity. With $\mathbf{Q}(\cdot)$ denoting the standard two-dimensional rotation matrix, the retained points are transformed into the centered, principal-axis-aligned local frame as
\begin{equation}
	\mathbf{p}_{i,m}
	=
	\mathbf{Q}(-\widehat{\psi}_i)
	\left(\mathbf{p}_{i,m}^{\mathrm{rad}}-\mathbf{c}_i\right),
	\qquad m\in\mathcal{K}_i,
	\label{eq:target_local_transform}
\end{equation}
and $\mathcal{P}_i^{+}=\{(\mathbf{p}_{i,m},a_{i,m})\}_{m\in\mathcal{K}_i}$. Thus, $R_i^{0}$ and $\theta_i^{0}$ come from ordinary radar range-gate and beam-pointing records, whereas the center and principal direction used for alignment are estimated directly from the reconstructed point set. No external target heading is assumed. Residual channel mismatch and noise may still introduce angular bias or dispersion; the calibration, denominator regularization, dominant-peak threshold, and main-cluster filtering jointly limit their influence. PCSP therefore treats the reconstruction as a compact physical prior rather than as precision geolocation.

For samples without the difference channel, $\Delta\theta_i(r)$ cannot be reliably estimated. PCSP then sets the angular deviation to zero and forms a range-consistent scattering skeleton along the radar line of sight. This form preserves the radial distribution, peak spacing, and amplitude allocation of dominant responses. Cross-range-dependent descriptor entries are omitted consistently for a range-only dataset, whereas the subsequent filtering, prototype construction, and recognition operations remain unchanged.

\subsubsection{Class-Level Scattering Prototype Construction}

For incremental recognition, $\mathcal{P}_i^{+}$ is converted into a compact physical descriptor:
\begin{equation}
	\begin{aligned}
		\mathbf{s}_i
		&=
		g(\mathcal{P}_i^{+}) \\
		&=
		\big[
		L_i,
		W_i,
		\gamma_i,
		\sin\psi_i,
		\cos\psi_i,
		\bar{\rho}_i, \\
		&\qquad
		\sigma_{\rho,i},
		E_{\mathrm{con},i},
		\tilde{N}_i
		\big]^{T}.
	\end{aligned}
	\label{eq:scattering_descriptor}
\end{equation}
where $g(\cdot)$ maps the filtered scattering-point set to a compact physical vector. It uses the amplitude-weighted center and covariance described above; target extent is obtained from the spans of the aligned points along the two principal axes. Here, $L_i$ and $W_i$ are the estimated length and width, $\gamma_i$ is the aspect ratio, $\psi_i=\widehat{\psi}_i-\theta_i^{0}$ is the principal orientation relative to the radar line of sight, $\bar{\rho}_i$ and $\sigma_{\rho,i}$ describe the radial energy distribution, $E_{\mathrm{con},i}$ measures the fraction of profile energy carried by the retained points, and $\tilde{N}_i$ is their normalized number. The sine--cosine encoding avoids discontinuity at the angular boundary. All dimensional quantities are consistently expressed in meters, and component-wise rescaling is intentionally omitted so that reconstructed target extent serves as the primary class-level physical cue, while the orientation and energy terms provide complementary information. Although individual observations need not yield identical descriptors, the class prototype below aggregates these attributes over all available training aspects. For range-only observations, $g(\cdot)$ retains the corresponding radial entries.

For each class $c$, a class-level scattering prototype is constructed as
\begin{equation}
	\mathbf{q}_c
	=
	\frac{1}{|\mathcal{I}_c|}
	\sum_{i\in\mathcal{I}_c}
	\frac{\mathbf{s}_i}{\|\mathbf{s}_i\|_2},
	\label{eq:class_scattering_prototype}
\end{equation}
where $\mathcal{I}_c$ denotes the training-sample index set of class $c$. The prototype $\mathbf{q}_c$ represents the typical scattering geometry of class $c$ in the reconstructed physical space. During class-incremental learning, old-class prototypes are stored as physical anchors, while new-class prototypes are estimated from the samples available at the current stage.

\subsubsection{Prototype-Constrained Incremental Recognition}

For notational simplicity, the stage index is omitted from the sample-wise features and scores in this subsection; all such quantities are evaluated using the current-stage model.

Given a complex observation $\mathbf{x}_i$, the feature extractor produces the recognition feature
\begin{equation}
	\mathbf{z}_i=f_{\Theta}(\widetilde{\mathbf{x}}_i),
	\label{eq:deep_feature_pcsp}
\end{equation}
where $f_{\Theta}(\cdot)$ is the current feature extractor and $\widetilde{\mathbf{x}}_i$ is the saliency-guided channel pair defined below. When saliency guidance is disabled, $\widetilde{\mathbf{x}}_i=\mathbf{x}_i$. The recognition feature is then projected into the physical-prototype space as
\begin{equation}
	\begin{aligned}
		h_{\Phi_t}(\mathbf{z}_i)
		&=h_{\Phi_{t-1}}(\mathbf{z}_i)
		+P_{\Delta\Phi_t}(\mathbf{z}_i),\quad t\geq1,\\
		\mathbf{u}_i
		&=\operatorname{norm}_2\!\left(h_{\Phi_t}(\mathbf{z}_i)\right),
	\end{aligned}
\label{eq:physical_projection_pcsp}
\end{equation}
where $\operatorname{norm}_2(\mathbf{v})=\mathbf{v}/\|\mathbf{v}\|_2$, $\Phi_t$ denotes the parameters of the projection mapping retained after stage $t$, and $\Delta\Phi_t$ denotes the parameters of the projection adapter introduced at that stage. The base projection $h_{\Phi_0}$ is learned at the base stage. At update $t$, $h_{\Phi_{t-1}}$ is frozen and only the compact projection adapter $P_{\Delta\Phi_t}$ is added, enabling new-class features to enter the common prototype space without overwriting the earlier mapping.

For a mini-batch $\mathcal{B}_t$ at stage $t$, the scattering prototype constraint is formulated as
\begin{equation}
	\mathcal{L}_{\mathrm{sp}}
	=
	-\frac{1}{|\mathcal{B}_t|}
	\sum_{i\in\mathcal{B}_t}
	\log
	\frac{
		\exp\left(
		\operatorname{sim}(\mathbf{u}_{i},\mathbf{q}_{y_i})/\tau_p
		\right)
	}{
		\sum_{c\in\mathcal{C}^{0:t}}
		\exp\left(
		\operatorname{sim}(\mathbf{u}_{i},\mathbf{q}_{c})/\tau_p
		\right)
	},
	\label{eq:sp_loss}
\end{equation}
where $|\mathcal{B}_t|$ is the mini-batch size, $\operatorname{sim}(\cdot,\cdot)$ denotes cosine similarity, $\tau_p$ is the temperature coefficient, and $\mathcal{C}^{0:t}$ is the set of all classes observed up to stage $t$. This loss pulls each sample toward the scattering prototype of its own class and pushes it away from prototypes of other observed classes.

The linear classifier score vector is subsequently corrected by the physical prototype similarity:
\begin{equation}
	\mathbf{o}^{\mathrm{net}}_{i}
	=
	\mathbf{o}^{\mathrm{lin}}_{i}
	+\lambda_{\mathrm{p}}
	\left[
	\operatorname{sim}(\mathbf{u}_{i},\mathbf{q}_c)
	\right]_{c\in\mathcal{C}^{0:t}},
	\label{eq:prototype_score}
\end{equation}
where $\mathbf{o}^{\mathrm{lin}}_{i}$ and $\mathbf{o}^{\mathrm{net}}_{i}$ are respectively the linear and PCSP-corrected score vectors of sample $i$ at the current stage, the bracketed term collects the similarities to all observed-class prototypes, and $\lambda_{\mathrm{p}}$ controls the prototype contribution. The final objective is
\begin{equation}
	\mathcal{L}
	=
	\mathcal{L}_{\mathrm{ce}}
	+
	\lambda_{\mathrm{sp}} \mathcal{L}_{\mathrm{sp}}
	+
	\lambda_{\mathrm{kd}} \mathcal{L}_{\mathrm{kd}},
	\label{eq:overall_loss_pcsp}
\end{equation}
where $\mathcal{L}_{\mathrm{ce}}$ is the cross-entropy loss evaluated from the PCSP-corrected scores in \eqref{eq:prototype_score}, $\mathcal{L}_{\mathrm{kd}}$ is the old-class distillation term defined in \Cref{subsec:lrae}, and $\lambda_{\mathrm{sp}}$ and $\lambda_{\mathrm{kd}}$ are their weighting coefficients.

\subsubsection{Sample-Level Physical Saliency Guidance}

The reconstructed scattering set also provides sample-level guidance for feature extraction. Because dominant scattering points correspond to physically meaningful range positions, a soft physical saliency mask is constructed as
\begin{equation}
	m_i(r)
	=
	\max_{(\mathbf{p}_{i,m},a_{i,m})\in\mathcal{P}_i^{+}}
	a_{i,m}
	\exp
	\left(
		-\frac{(r-\Pi_{i,r}(\mathbf{p}_{i,m}))^2}{2\sigma_{\mathrm{sal}}^2}
	\right),
	\label{eq:physical_mask}
\end{equation}
where $\Pi_{i,r}(\cdot)$ denotes the sample-specific inverse mapping from the aligned physical plane back to the HRRP range index, and $\sigma_{\mathrm{sal}}$ controls the smoothness of the saliency response.

The mask is normalized as $\bar{m}_i(r)=m_i(r)/(\max_r m_i(r)+\epsilon_{\mathrm{sal}})$, where $\epsilon_{\mathrm{sal}}$ is a numerical-stability constant, and is used to reweight, rather than discard, the original response:
\begin{equation}
	\widetilde{x}_{i,k}(r)
	=
	x_{i,k}(r)\bigl[1+\bar{m}_i(r)\bigr],
	\qquad k\in\{\Sigma,\Delta\}.
	\label{eq:saliency_reweighting}
\end{equation}
The feature extractor in \eqref{eq:deep_feature_pcsp} receives $\widetilde{\mathbf{x}}_i$ when saliency guidance is enabled. The same real-valued weight is applied to both complex channels and therefore preserves their phases. Thus, PCSP emphasizes physically supported range cells without discarding any component of the original observation.
\subsection{PSD-Guided Scattering-Component Inference Calibration}
\label{subsec:psd_scic}

Reflector groups can be confused with ships because a few strong responses may occupy most of the profile energy and produce similar range-domain peak patterns. PSD-SCIC addresses this ambiguity through a fixed scattering-mechanism verifier that operates only at inference. It is trained independently of the radar recognition network and therefore does not alter the incremental feature-learning process.

The primitive scatterer dictionary (PSD) describes point, line or edge, surface or curved-surface, and multiple-bounce scattering structures~\cite{Zhang2023}. These primitives provide reference mechanisms for generating standard local range-profile responses. For clarity, the dictionary is written explicitly as
\begin{equation}
	\mathrm{PSD\ Dictionary}
	=
	\left\{\mathrm{Primitive}_{\nu}\right\}_{\nu=1}^{N_{\mathrm{psd}}},
\end{equation}
where $\mathrm{Primitive}_{\nu}$ denotes the $\nu$-th primitive type and $N_{\mathrm{psd}}$ is the number of primitive types. The 13 primitive types specified by the PSD are used in the implementation.

The PSD forward model generates local response segments under different component parameters and observation configurations. These simulated segments form the training set of a scattering-component classifier:
\begin{equation}
	\mathcal{H}_{\boldsymbol{\varpi}}(\boldsymbol{\upsilon})
	=
	\boldsymbol{\pi}^{\mathrm{psd}},
\end{equation}
where $\boldsymbol{\varpi}$ denotes the parameters of the PSD classifier, $\boldsymbol{\upsilon}$ denotes a local response segment, and $\boldsymbol{\pi}^{\mathrm{psd}}\in\mathbb{R}^{N_{\mathrm{psd}}}$ is the predicted PSD component distribution. The classifier learns to describe a local response by its scattering mechanism rather than by the target category. After this pretraining stage, $\mathcal{H}_{\boldsymbol{\varpi}}(\cdot)$ is fixed and is not updated on the target recognition dataset. This design avoids coupling the PSD calibration module with the incremental training process and allows it to serve as an external physical reference.

For a radar observation $\mathbf{x}_i$, local scattering segments are extracted around its dominant peaks. Let $\boldsymbol{\upsilon}_{i,\ell}$ denote the local segment associated with the $\ell$-th dominant peak. The fixed PSD classifier then produces its scattering-component distribution as
\begin{equation}
	\boldsymbol{\pi}^{\mathrm{psd}}_{i,\ell}
	=
	\mathcal{H}_{\boldsymbol{\varpi}}(\boldsymbol{\upsilon}_{i,\ell}).
\end{equation}
Thus, each dominant peak is assigned a soft PSD component description. Compared with directly using peak amplitude or position, this description provides a more explicit indication of whether the local response is closer to a surface, edge, point-like, or multiple-bounce scattering mechanism.

For each target class, a PSD mechanism template is computed from the training samples. Specifically, all dominant-peak segments of class $c$ are passed through the fixed PSD classifier, and their component distributions are averaged as
\begin{equation}
	\boldsymbol{\chi}^{\mathrm{psd}}_{c}
	=
	\frac{1}{Z_c^{\mathrm{psd}}}
	\sum_{i\in\mathcal{I}^{\mathrm{psd}}_c}
	\sum_{\ell=1}^{J_i}
	e_{i,\ell}
	\boldsymbol{\pi}^{\mathrm{psd}}_{i,\ell},
\end{equation}
where $\mathcal{I}^{\mathrm{psd}}_c$ is the training-sample index set of class $c$ for PSD template construction, $J_i$ is the number of selected dominant peaks, and $e_{i,\ell}$ is the energy of the local segment normalized within sample $i$. The normalization factor is
\begin{equation}
	Z_c^{\mathrm{psd}}=
	\sum_{i\in\mathcal{I}^{\mathrm{psd}}_c}
	\sum_{\ell=1}^{J_i}
	e_{i,\ell}.
\end{equation}
The template $\boldsymbol{\chi}^{\mathrm{psd}}_{c}$ describes the typical composition of PSD scattering components for class $c$. It is a fixed mechanism-level reference for inference calibration, rather than a trainable feature prototype. In particular, reflector groups tend to have a more concentrated mechanism composition dominated by specular or multiple-bounce responses, whereas ship targets usually exhibit a richer mixture of scattering components caused by decks, sides, edges, curved surfaces, and local structures.

Although $\mathcal{H}_{\boldsymbol{\varpi}}$ is pretrained using simulated primitive-scatterer responses, PSD-SCIC does not perform direct synthetic-to-real target classification. The simulated responses only establish a target-independent coordinate system over the predefined scattering mechanisms; they are not used to learn ship or reflector-group labels. For each radar dataset, $\boldsymbol{\chi}^{\mathrm{psd}}_{c}$ is constructed from the training observations of that dataset after their local scattering segments are mapped into the fixed PSD space. A test observation is therefore compared with class templates from the same observation domain, rather than directly with simulated primitive responses. The calibration relies on relative mechanism consistency and does not require every local component to be assigned an exactly correct primitive label. Energy-weighted aggregation over multiple dominant components further reduces the influence of individual mapping deviations.

Moreover, PSD-SCIC serves only as a conservative inference-time verifier. It is invoked for uncertain ship--reflector decisions and modifies only the two ambiguous candidate scores; the PSD classifier neither replaces the recognition network nor independently determines the target category. Specifically, the recognition network first produces the PCSP-corrected score vector $\mathbf{o}^{\mathrm{net}}_{i}$ in \eqref{eq:prototype_score} and the corresponding probability vector $\mathbf{p}^{\mathrm{net}}_{i}=\operatorname{softmax}(\mathbf{o}^{\mathrm{net}}_{i})$. Calibration is triggered when the two classes with the highest probabilities contain one ship and one reflector group and their probability difference is below an ambiguity threshold $\delta_{\mathrm{amb}}$. Otherwise, the original network scores are retained. This selective trigger prevents unnecessary changes to confident predictions and to decisions that do not involve reflector interference.

For a candidate class $c$, the PSD matching score is computed by comparing the local PSD component distributions of the test sample with the class template:
\begin{equation}
	\Gamma^{\mathrm{psd}}_{i,c}
	=
	\frac{1}{\sum_{\ell=1}^{J_i} e_{i,\ell}}
	\sum_{\ell=1}^{J_i}
	e_{i,\ell}
		\operatorname{sim}
		\left(
		\boldsymbol{\pi}^{\mathrm{psd}}_{i,\ell},
		\boldsymbol{\chi}^{\mathrm{psd}}_{c}
		\right).
\end{equation}
Only the two ambiguous candidate classes are recalibrated:
\begin{equation}
	[\mathbf{o}^{\mathrm{cal}}_{i}]_c
	=
	[\mathbf{o}^{\mathrm{net}}_{i}]_c
	+
	\mu_{\mathrm{sc}}\Gamma^{\mathrm{psd}}_{i,c},
\end{equation}
where $\mu_{\mathrm{sc}}$ controls the strength of scattering-component calibration; $\operatorname{sim}(\cdot,\cdot)$ is the cosine similarity defined in \eqref{eq:sp_loss}. Scores of the remaining classes are unchanged.

The calibrated score combines the discriminative evidence learned from radar observations with the mechanism-level evidence provided by PSD. By checking whether the dominant local responses are more consistent with ship-like component mixtures or reflector-like scattering mechanisms, PSD-SCIC improves robustness to reflector-induced interference without modifying the learned representation.

\subsection{Lightweight Residual Adapter Expansion}
\label{subsec:lrae}

Dynamic expansion can isolate new-class learning from old representations, but duplicating a complete feature extractor at every stage causes rapid model growth. LRAE instead shares one HRRP backbone and adds only a compact residual adapter for each new stage.

Within the complete model, let $\mathbf{h}_i\in\mathbb{R}^{C}$ denote the feature of sample $i$ produced by the shared backbone before stage-specific adaptation, where $C$ is its dimension. For an incremental stage, LRAE introduces a bottleneck residual adapter instead of a new backbone:
\begin{equation}
	A_t(\mathbf{h}_i)
	=
	\mathbf{W}_{\mathrm{up},t}
	\varphi
	\left(
	\mathbf{W}_{\mathrm{down},t}\mathbf{h}_i
	\right),
\end{equation}
where $\mathbf{W}_{\mathrm{down},t}\in\mathbb{R}^{b_{\mathrm{ad}}\times C}$ and $\mathbf{W}_{\mathrm{up},t}\in\mathbb{R}^{C\times b_{\mathrm{ad}}}$ are the stage-$t$ down- and up-projection matrices, respectively, and $\varphi(\cdot)$ is the nonlinear activation. The bottleneck dimension $b_{\mathrm{ad}}\ll C$ limits the newly introduced parameters to approximately $2Cb_{\mathrm{ad}}$.

At stage $t$, the adapter bank $\{A_j\}_{j=1}^{t}$ contains the residual corrections learned at all observed stages. Their contributions are fused using normalized learnable weights
\begin{equation}
	\omega_{t,j}
	=
	\frac{\exp(\zeta_{t,j})}
	{\sum_{j'=1}^{t}\exp(\zeta_{t,j'})},
	\qquad j=1,\ldots,t,
	\label{eq:lrae_adapter_weight}
\end{equation}
and the expanded recognition feature is
\begin{equation}
	\mathbf{h}_{i,t}^{\prime}
	=
	\mathbf{h}_i
	+
	\sum_{j=1}^{t}
	\omega_{t,j}A_j(\mathbf{h}_i),
	\label{eq:lrae_adapter_bank}
\end{equation}
where $\zeta_{t,j}$ is a learnable fusion logit and the softmax normalization prevents the accumulated residuals from growing with the number of stages. Historical adapters remain active, while the model learns their relative contributions at the current stage. Accordingly, the feature $\mathbf{z}_i$ used in PCSP equals $\mathbf{h}_i$ at the base stage and $\mathbf{h}_{i,t}^{\prime}$ at incremental stage $t$.

This residual form has two advantages. First, the shared backbone feature $\mathbf{h}_i$ is preserved as the common radar representation, which contains general range-domain scattering patterns learned from previous data. Second, the adapter bank provides stage-specific residual corrections with only a small number of additional parameters. Therefore, the model can expand its representation capacity for new classes while maintaining previously learned residual knowledge.

At the base stage, the shared backbone, initial classifier, and base projection $h_{\Phi_0}$ are jointly optimized, and no residual adapter is required. At update $t\geq1$, the backbone, historical adapters, the previous PCSP projection mapping, and old-class classifier rows remain fixed. Only $A_t$, the fusion logits $\{\zeta_{t,j}\}_{j=1}^{t}$, the new classifier rows, and the PCSP projection adapter $P_{\Delta\Phi_t}$ are optimized. The expanded linear classifier produces
\begin{equation}
	\mathbf{o}^{\mathrm{lin}}_{i,t}
	=
	\mathbf{W}_{\mathrm{cls},t}\mathbf{h}_{i,t}^{\prime},
\end{equation}
where $\mathbf{W}_{\mathrm{cls},t}$ denotes the classifier after expansion to stage $t$; $\mathbf{o}^{\mathrm{lin}}_{i,t}$ corresponds to $\mathbf{o}^{\mathrm{lin}}_i$ in PCSP, where the stage index is omitted. The resulting vector is corrected by PCSP according to \eqref{eq:prototype_score} before evaluating the classification loss. Thus, LRAE supplies the expandable representation and linear evidence, whereas PCSP provides the subsequent physical score correction.

The incremental regularizer preserves old-class responses by knowledge distillation on the PCSP-corrected network scores. Let $\mathcal{S}_{T}(\mathbf{v})=\operatorname{softmax}(\mathbf{v}/T)$ denote temperature-scaled softmax. Over the same mini-batch, the distillation term is
\begin{equation}
	\mathcal{L}_{\mathrm{kd}}
	=
	\frac{T_{\mathrm{kd}}^{2}}{|\mathcal{B}_t|}
	\sum_{i\in\mathcal{B}_t}
	\operatorname{KL}\!\left(
	\mathcal{S}_{T_{\mathrm{kd}}}(\mathbf{o}^{\mathrm{net,old}}_{i,t-1})
	\,\middle\|\,
	\mathcal{S}_{T_{\mathrm{kd}}}(\mathbf{o}^{\mathrm{net,old}}_{i,t})
	\right),
	\label{eq:lrae_distillation}
\end{equation}
where $\mathbf{o}^{\mathrm{net,old}}_{i,t-1}$ and $\mathbf{o}^{\mathrm{net,old}}_{i,t}$ are the previous- and current-model network scores restricted to old classes, $T_{\mathrm{kd}}$ is the distillation temperature, and $\operatorname{KL}(\cdot\|\cdot)$ denotes Kullback--Leibler divergence. Its contribution is controlled by $\lambda_{\mathrm{kd}}$ in \eqref{eq:overall_loss_pcsp}. PSD-SCIC is applied only after this training process and therefore does not enter either loss term.

Compared with full-branch expansion, LRAE provides a lighter way to increase the model capacity. The shared backbone maintains a stable radar representation, while each incremental stage only contributes a compact residual adapter. Therefore, the model can adapt to new classes without linearly duplicating large feature extractors, achieving a better balance among plasticity, stability, and parameter efficiency.
\section{Experiments}
\subsection{Datasets}
\begin{figure}[!t]
	\centering
	\includegraphics[width=0.95\columnwidth]{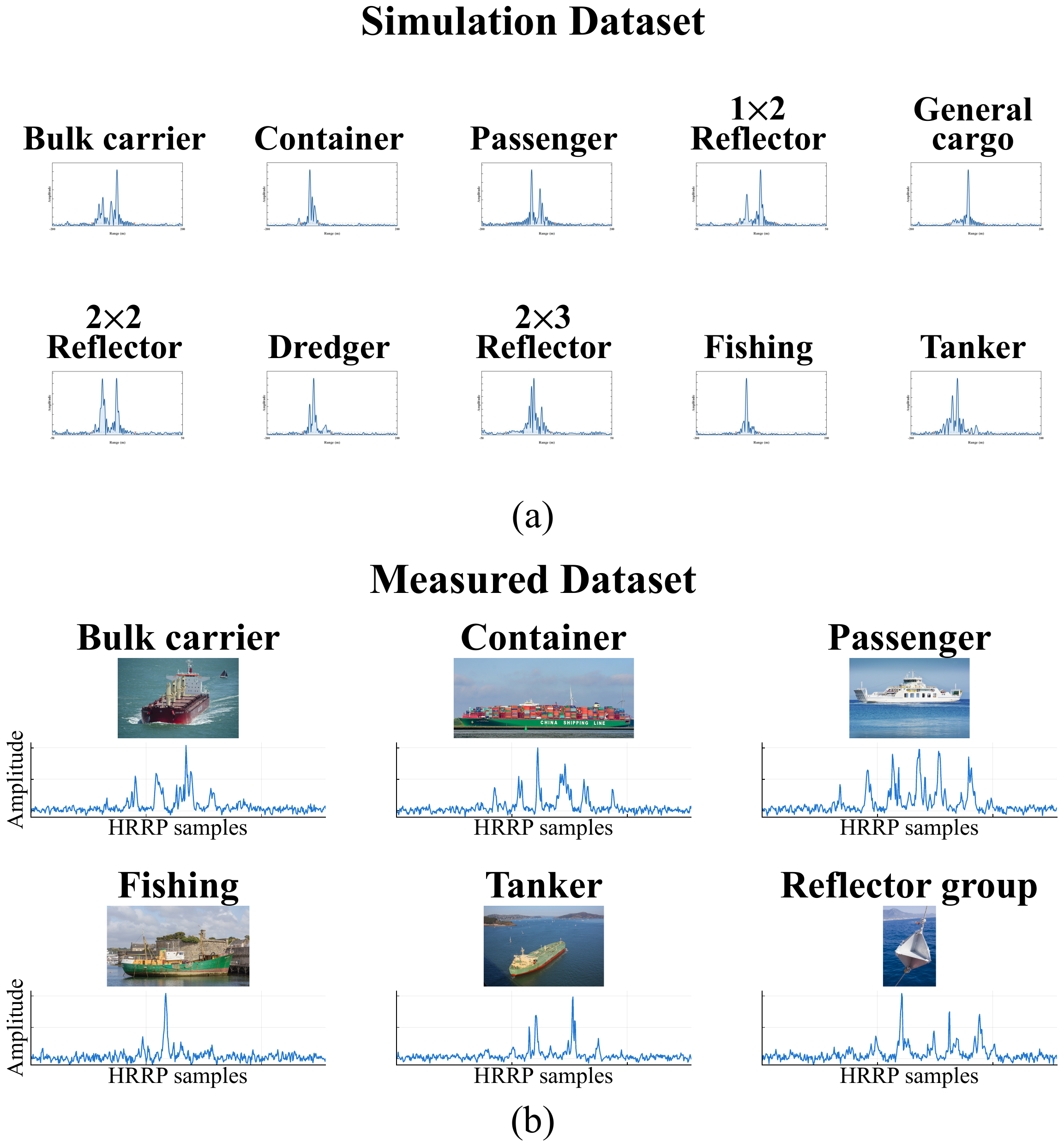}
	\caption{Representative samples from the simulation and measured datasets: (a) HRRP profiles of the ten simulated target classes, including ships and reflector groups; and (b) measured target images with their corresponding HRRP profiles.}
	\label{fig:simulation_measured_hrrp}
\end{figure}

Experiments are conducted using a self-constructed simulation dataset and a measured HRRP dataset. Both datasets provide synchronized complex sum-channel and azimuth-difference-channel responses for each observation. Representative profiles from both datasets are shown in \Cref{fig:simulation_measured_hrrp}.
\subsubsection{Self-Constructed Simulation Dataset}
The simulation dataset is generated from electromagnetic models of seven ship targets and three reflector-group configurations, yielding ten target classes. The simulated radar operates in the X-band with a center frequency of 10~GHz and a bandwidth of 500~MHz, corresponding to a nominal range resolution of 0.30~m. The simulated observations cover pitch angles of $0^{\circ}$--$90^{\circ}$ and azimuth angles of $0^{\circ}$--$360^{\circ}$. Each class contains 4050 valid HRRPs, and each sample consists of 1024 range cells. The class composition, physical target sizes, and training--test partitions are summarized in \Cref{tab:datasets}.
\begin{table}[bp]
	\caption{Simulation and Measured HRRP Datasets}
	\centering
	\resizebox{\columnwidth}{!}{
		\begin{tabular}{lcccc}
			\toprule
			\textbf{Dataset} & \textbf{Target type} & \textbf{Length \& width (m)} & \textbf{Training} & \textbf{Test} \\
			\midrule
			\textbf{Simulation dataset} & & & & \\
			& Bulk carrier & (152.14, 16.07) & 3240 & 810 \\
			& Container & (139.37, 23.37) & 3240 & 810 \\
			& Passenger & (178.82, 26.01) & 3240 & 810 \\
			& General cargo & (121.50, 21.72) & 3240 & 810 \\
			& Dredger & (79.52, 15.42) & 3240 & 810 \\
			& Fishing & (46.85, 8.60) & 3240 & 810 \\
			& Tanker & (168.55, 18.44) & 3240 & 810 \\
			& $1\times2$ reflector & (44.40, 7.20) & 3240 & 810 \\
			& $2\times2$ reflector & (44.40, 24.40) & 3240 & 810 \\
			& $2\times3$ reflector & (81.60, 24.40) & 3240 & 810 \\
			\midrule
			\textbf{Measured dataset} & & & & \\
			& Bulk carrier & (156.68, 18.06) & 8631 & 2158 \\
			& Container & (225.00, 32.20) & 7463 & 1866 \\
			& Passenger & (183.40, 25.00) & 4679 & 1170 \\
			& Fishing & (35.56, 9.30) & 12676 & 3169 \\
			& Tanker & (248.57, 44.14) & 6780 & 1695 \\
			& Reflector group & (48.20, 26.20) & 4855 & 1214 \\
			\bottomrule
		\end{tabular}
	}
	\label{tab:datasets}
\end{table}

\subsubsection{The Measured Dataset}
The measured dataset contains six target classes, and each synchronized complex channel pair is annotated with its class name. The radar operates at a center frequency of 10~GHz and a bandwidth of 120~MHz, corresponding to a nominal range resolution of 1.25~m. The measurements cover pitch angles of $0^{\circ}$--$25^{\circ}$ and azimuth angles of $0^{\circ}$--$360^{\circ}$. All valid observations are retained, with class-dependent sample counts reported together with the target sizes and data partitions in \Cref{tab:datasets}.

For both datasets, the training and test sets are partitioned by disjoint azimuth intervals rather than by randomly assigning individual samples. Samples from a given azimuth interval appear in only one partition, which prevents adjacent-aspect observations of the same target from being shared between training and testing. The resulting sample counts are listed in \Cref{tab:datasets}.
\begin{figure*}[!b]
	\centering
	\includegraphics[width=\textwidth]{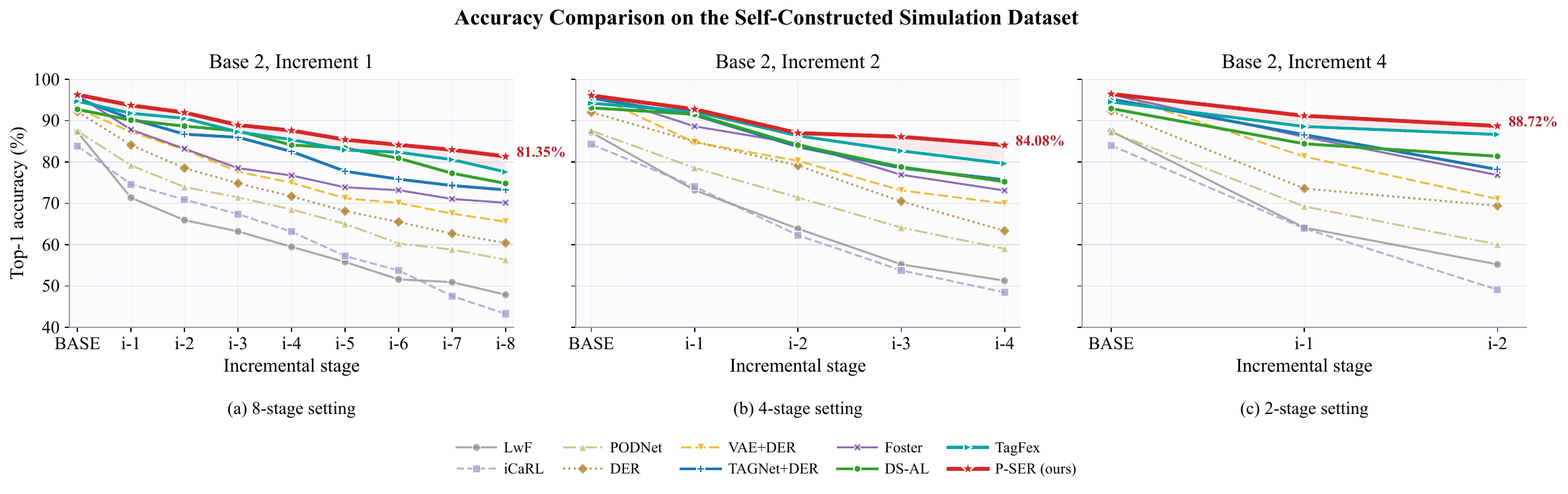}
	\caption{Accuracy comparison of different incremental learning methods across incremental stages on the self-constructed simulation dataset.}
	\label{fig:simulation_accuracy_stages}
\end{figure*}
\subsection{Experimental Settings}\label{sec:experimental_settings}
All protocols begin with two base classes. For the ten-class simulation dataset, adding 1, 2, or 4 classes per update yields the 8-stage, 4-stage, and 2-stage settings. The class order is bulk carrier, container, passenger, $1\times2$ reflector, general cargo, $2\times2$ reflector, dredger, $2\times3$ reflector, fishing, and tanker. For the six-class measured dataset, adding 1 or 2 classes per update yields the 4-stage and 2-stage settings, with the order bulk carrier, container, passenger, fishing, tanker, and reflector group. All methods use identical class orders and data partitions.

All methods use the same one-dimensional ResNet-18. Before entering the real-valued backbone, the complex observation $\mathbf{x}_i$ defined in Section~III is rearranged into the four-channel real-valued tensor
\begin{equation}
	\mathbf{x}_i^{\mathrm{RI}}=
	[\operatorname{Re}(\mathbf{S}_i),\operatorname{Im}(\mathbf{S}_i),
	\operatorname{Re}(\mathbf{D}_i),\operatorname{Im}(\mathbf{D}_i)]^{T}
	\in\mathbb{R}^{4\times N_r}.
\end{equation}
All compared methods receive the same complex sum- and azimuth-difference-channel observations. P-SER additionally exploits routinely available radar-side geometry and calibration information, including the beam-pointing angle $\theta_i^{0}$, the range-gate reference $R_i^{0}$, the channel responses $G_{\Sigma}$ and $G_{\Delta}$, and the monopulse slope $\kappa_{\mathrm{mp}}$, to construct its physical prior. The former two are acquisition metadata recorded for each observation, whereas the latter quantities are target-independent calibration parameters fixed for a given radar configuration. No target heading, physical dimension, class label, or additional target measurement is required at inference. The backbone retains the ResNet-18 block configuration $[2,2,2,2]$ and widths $[64,128,256,512]$, with one-dimensional operations, a four-channel seven-tap stem, and global average pooling that yields $C=512$. Adam is used for 200 epochs at each stage with an initial learning rate of $10^{-3}$. The remaining settings are $b_{\mathrm{ad}}=32$, $\lambda_{\mathrm{kd}}=1.0$, $T_{\mathrm{kd}}=2.0$, $\lambda_{\mathrm{sp}}=0.10$, $\lambda_{\mathrm{p}}=0.10$, $\tau_p=0.07$, and $\mu_{\mathrm{sc}}=0.20$.

Candidate peaks are local maxima within seven-cell neighborhoods whose amplitudes exceed $10\%$ of the profile maximum, giving $\beta=0.10$. The saliency width is $\sigma_{\mathrm{sal}}=3$ cells, and numerical-stability constants are $10^{-8}$.

The fixed PSD classifier processes the real and imaginary parts of an 11-cell local sum-channel segment. Three one-dimensional convolutional layers with 32, 64, and 128 channels, each followed by batch normalization and ReLU, are connected to global average pooling and a 13-way classifier. The PSD forward model generates 3000 segments per primitive, or 39000 in total, using amplitude scaling in $[0.8,1.2]$, peak shifts of at most two cells, and Gaussian noise from 10 to 30~dB. The data are split 8:1:1, and the classifier is pretrained for 100 epochs with Adam, a batch size of 256, a learning rate of $10^{-3}$, and a weight decay of $10^{-4}$ before being fixed. PSD-SCIC uses $\delta_{\mathrm{amb}}=0.15$. Experiments are implemented in PyTorch on an NVIDIA RTX~4090 GPU.

Let $\mathcal{A}_b$ denote the Top-1 accuracy after update $b$, where $b=0$ is the base stage and $B$ is the final update. Performance over the complete sequence is summarized by
\begin{equation}
	\bar{\mathcal{A}}
	=
	\frac{1}{B+1}
	\sum_{b=0}^{B}\mathcal{A}_b.
	\label{eq:average_accuracy}
\end{equation}
where $\mathcal{A}_B$ and $\bar{\mathcal{A}}$ measure final-stage retention and average incremental performance, respectively.

The baselines include the distillation methods LwF~\cite{Li2017}, iCaRL~\cite{Rebuffi2017}, and PODNet~\cite{Douillard2020}; the expansion methods DER~\cite{Yan2021} and FOSTER~\cite{Wang2022}; DER equipped with VAE~\cite{Du2019} or TAGNet~\cite{Liu2025}; the exemplar-free DS-AL~\cite{Zhuang2024}; and TagFex~\cite{Zheng2025}. Only method-specific components differ; the backbone, four-channel input, class order, and data split are identical. Reported exemplar counts are per-old-class raw-observation budgets. P-SER stores no historical raw profiles, retaining only one PCSP prototype and one PSD template per observed class.
\subsection{Experiments on the Self-Constructed Simulation Dataset}

Following the settings in \Cref{sec:experimental_settings}, class-incremental learning experiments are conducted on the self-constructed simulation dataset. The dataset contains ten classes comprising ship targets and reflector-group configurations. To evaluate the adaptability of different methods under different incremental granularities, three independent protocols are considered: the 8-stage, 4-stage, and 2-stage settings.

\Cref{fig:simulation_accuracy_stages} presents the stage-wise recognition accuracy curves under the three protocols. Overall, because complete historical training data are unavailable during incremental updates, all CIL methods suffer from different degrees of catastrophic forgetting. LwF mainly constrains the output of the updated model through knowledge distillation and therefore shows limited ability to preserve old scattering structures. Although iCaRL additionally retains exemplars, its representation still degrades markedly as more classes arrive. PODNet introduces a finer-grained pooled distillation strategy and performs better than LwF and iCaRL, but it remains insufficient for handling strong aspect sensitivity and peak-structure variation.

In contrast, DER and FOSTER, as representative dynamic expansion methods, consistently outperform conventional distillation-based approaches. This result indicates that allocating additional representation capacity is beneficial for learning newly introduced radar target classes. When the range-profile-oriented VAE and TAGNet models are combined with DER, recognition performance improves further, demonstrating that domain-specific feature modeling can reduce representation drift in incremental recognition. DS-AL also remains competitive without old-class exemplars because its analytic mapping and compensation stream partially recover historical feature distributions. TagFex learns task-agnostic features to reduce the risk of overfitting to short-sighted peak patterns or spurious correlations when new-class samples are limited, and consequently achieves strong results across all three protocols.

Among all compared methods, P-SER achieves the highest recognition accuracy under the 8-stage, 4-stage, and 2-stage settings. In the 8-stage experiment, its accuracy curve remains consistently above those of the other methods as the incremental process proceeds, and its final accuracy of 81.35\% represents an absolute improvement of 3.75\% over the second-best result. In the 4-stage and 2-stage settings, P-SER likewise obtains the highest final accuracies of 84.08\% and 88.72\%, respectively. More importantly, its accuracy curves exhibit a smaller degradation slope during later updates. This behavior indicates that the proposed method not only adapts to the scattering characteristics of the radar observations but also effectively mitigates old-class forgetting during continuous class expansion.

A more comprehensive comparison is provided in \Cref{tab:comparison}, which reports the number of old-class exemplars, model parameters, final accuracy $\mathcal{A}_{B}$, and average accuracy $\bar{\mathcal{A}}$. Although distillation-based methods such as LwF, iCaRL, and PODNet use relatively compact models and may therefore be suitable for deployment-constrained scenarios, their final and average accuracies remain limited. DS-AL does not require old-class exemplars, but its analytic mappings and correlation matrices introduce considerable model storage overhead. By contrast, P-SER achieves the highest average accuracy without retaining any old-class exemplars. In the 8-stage setting, P-SER uses only 19.2\% of the parameters of TagFex while improving the average accuracy to 88.05\%, an absolute gain of 2.14\% over the second-best result. In the 4-stage and 2-stage settings, P-SER also obtains the highest average accuracies of 89.21\% and 92.13\%, respectively. These results demonstrate that P-SER achieves a favorable balance among recognition accuracy, storage efficiency, and incremental stability.
\begin{table*}[t]
	\caption{Performance Comparison on the Self-Constructed Simulation Dataset}
	\label{tab:comparison}
	\centering
	\scriptsize
	\renewcommand{\arraystretch}{1.0}
	\resizebox{0.9\textwidth}{!}{
		\begin{tabular}{l@{\hspace{3pt}}c@{\hspace{3pt}}c@{\hspace{3pt}}c@{\hspace{3pt}}c@{\hspace{3pt}}c@{\hspace{3pt}}c@{\hspace{3pt}}c@{\hspace{3pt}}c@{\hspace{3pt}}c@{\hspace{3pt}}c}  
			\toprule
			\multirow{2}{*}{\textbf{Method}} 
			& \multirow{2}{*}{\shortstack{\textbf{Exemplars}\\\textbf{per class}}}
			& \multicolumn{3}{c}{\textbf{8-stage}} 
			& \multicolumn{3}{c}{\textbf{4-stage}} 
			& \multicolumn{3}{c}{\textbf{2-stage}} \\  
			\cmidrule(lr){3-5} \cmidrule(lr){6-8} \cmidrule(lr){9-11}  
			& 
			& \textbf{Params. (M)} & \textbf{$\mathcal{A}_{B}$ (\%)} & \textbf{$\bar{\mathcal{A}}$ (\%)} 
			& \textbf{Params. (M)} & \textbf{$\mathcal{A}_{B}$ (\%)} & \textbf{$\bar{\mathcal{A}}$ (\%)} 
			& \textbf{Params. (M)} & \textbf{$\mathcal{A}_{B}$ (\%)} & \textbf{$\bar{\mathcal{A}}$ (\%)} \\  
			\midrule
			LwF\cite{Li2017} & 0 & 11.17 & 47.89 & 61.50 & 11.17 & 51.29 & 66.15 & 11.17 & 55.23 & 68.93 \\
			iCaRL\cite{Rebuffi2017} & 150 & 11.18 & 43.28 & 62.42 & 11.18 & 48.45 & 64.57 & 11.18 & 49.11 & 65.69 \\
			PODNet\cite{Douillard2020} & 150 & 11.18 & 56.34 & 69.01 & 11.18 & 59.03 & 72.15 & 11.18 & 60.05 & 72.18 \\
			DER\cite{Yan2021} & 150 & 89.39 & 60.41 & 73.12 & 44.71 & 63.37 & 77.98 & 22.37 & 69.41 & 78.45 \\
			VAE+DER\cite{Du2019,Yan2021} & 150 & 89.44 & 65.55 & 76.69 & 44.74 & 69.97 & 80.91 & 22.40 & 71.06 & 83.03 \\
			TAGNet+DER\cite{Liu2025,Yan2021} & 100 & 89.41 & 73.26 & 82.47 & 44.72 & 75.69 & 84.96 & 22.38 & 78.20 & 86.68 \\
			FOSTER\cite{Wang2022} & 150 & 71.08 & 70.14 & 78.96 & 35.55 & 73.11 & 83.97 & 17.78 & 76.84 & 86.42 \\
			DS-AL\cite{Zhuang2024} & 0 & 87.60 & 74.82 & 84.39 & 87.60 & 75.24 & 84.56 & 87.60 & 81.39 & 86.24 \\
			TagFex\cite{Zheng2025} & 100 & 63.22 & 77.60 & 85.91 & 35.94 & 79.62 & 87.01 & 18.71 & 86.65 & 89.93 \\
			P-SER & 0 & 12.12 & \textbf{81.35} & \textbf{88.05} & 11.99 & \textbf{84.08} & \textbf{89.21} & 11.92 & \textbf{88.72} & \textbf{92.13} \\
			\bottomrule
		\end{tabular}
	}
\end{table*}

\begin{figure*}[t]
	\centering
	\includegraphics[width=0.8\textwidth]{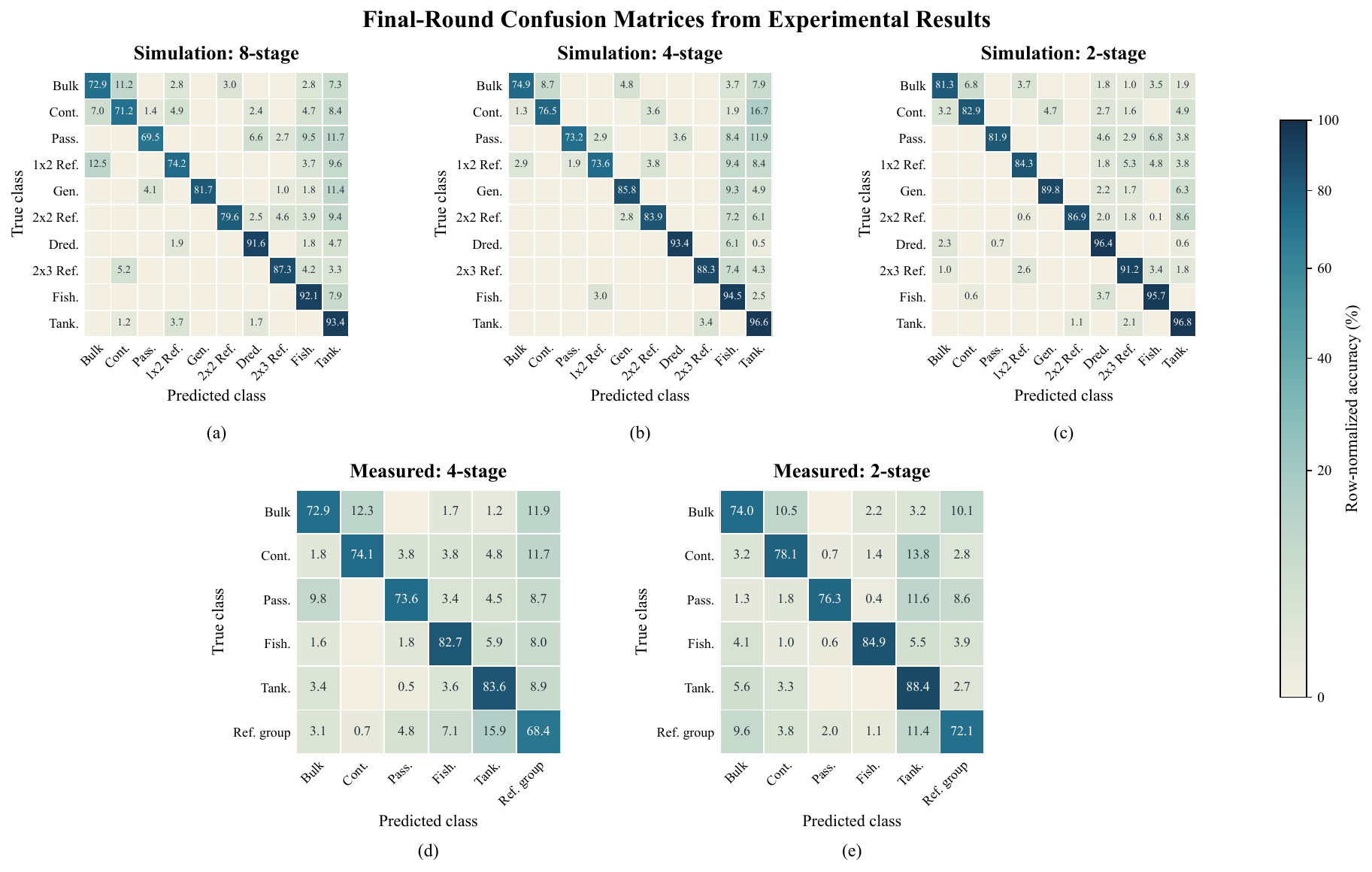}
	\caption{Final-round confusion matrices of P-SER. The top row reports the simulation results under the (a) 8-stage, (b) 4-stage, and (c) 2-stage settings; the bottom row reports the measured results under the (d) 4-stage and (e) 2-stage settings.}
	\label{fig:final_round_confusion_matrices}
\end{figure*}
\subsection{Experiments on the Measured Dataset}

The proposed method is further evaluated on the measured radar dataset. Compared with the simulation data, the measured observations are affected by target-aspect variation, environmental clutter, measurement errors, and class-dependent sample imbalance, which lead to less stable peak positions and amplitudes. The measured experiment therefore provides a more direct assessment of the incremental adaptation capability and practical robustness of the compared methods. For the six measured target classes, the 4-stage and 2-stage protocols described in \Cref{sec:experimental_settings} are adopted.

\begin{figure}[t]
	\centering
	\includegraphics[width=\columnwidth]{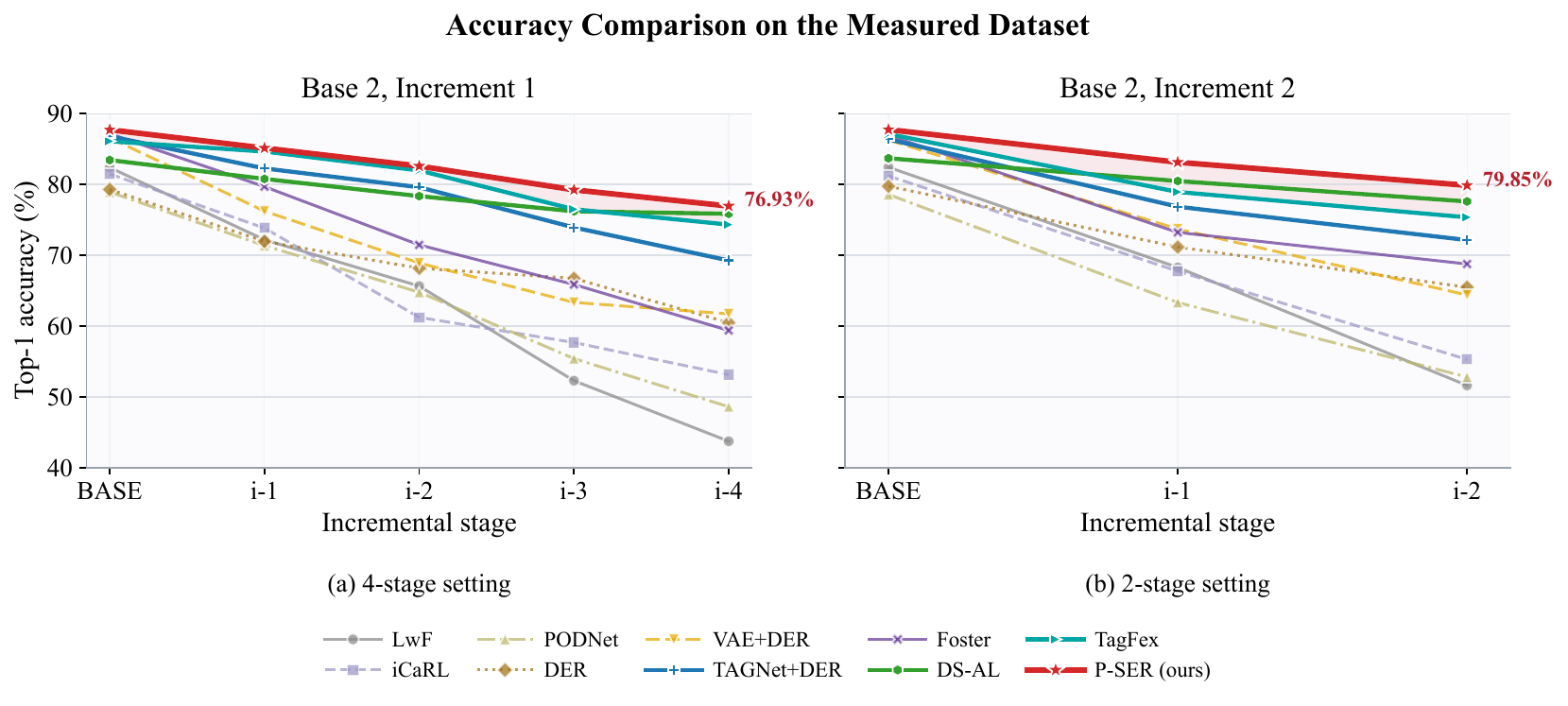}
	\caption{Accuracy comparison of different incremental learning methods across incremental stages on the measured dataset.}
	\label{fig:measured_accuracy_stages}
\end{figure}

\Cref{fig:measured_accuracy_stages} shows the stage-wise recognition accuracy of the different methods. The overall accuracies are lower than those obtained on the simulation dataset, and fluctuations among stages are more pronounced. Distillation-based methods, including LwF, iCaRL, and PODNet, continue to show limited performance, indicating that constraining only output responses or intermediate features is insufficient to preserve the aspect-dependent scattering structures of old classes. DER and FOSTER improve new-class learning through dynamic expansion and generally outperform conventional distillation methods. Incorporating VAE or TAGNet into DER provides a further improvement, confirming that range-profile-oriented feature modeling remains beneficial under real measurements. DS-AL and TagFex also achieve competitive results, reflecting the effectiveness of exemplar-free analytic learning and task-agnostic feature transfer in measured radar tasks.

P-SER achieves the best performance under both measured protocols. In the 4-stage experiment, it obtains a final accuracy of 76.93\% and an average accuracy of 82.30\%. In the 2-stage experiment, these metrics reach 79.85\% and 83.56\%, respectively. The corresponding accuracy curves also exhibit a smaller decline during later updates. This result suggests that the physical-consistency scattering prior and lightweight adapter expansion jointly reduce old-class representation drift, while PSD-guided calibration improves discrimination when reflector responses resemble ship measurements.

The bottom row of \Cref{fig:final_round_confusion_matrices} further reports the final-round confusion matrices on the measured dataset. Under the adopted incremental class order, many misclassifications of earlier ship classes are directed toward classes introduced in later stages, which is consistent with the feature drift and catastrophic forgetting commonly observed in CIL. Nevertheless, P-SER maintains stable discrimination between ships and the reflector group. Reflector-related errors remain comparatively limited despite the presence of strong local peaks, indicating that PSD-SCIC provides an effective mechanism-level check for ambiguous ship--reflector predictions. Meanwhile, PCSP helps preserve class-level geometric scattering differences among ship categories. The confusion matrices therefore support the overall interference robustness of P-SER, while the independent contribution of each component is quantified by the ablation experiments below.

\Cref{tab:comparison2} compares the methods in terms of model size, the number of old-class exemplars, final accuracy, and average accuracy. Although LwF, iCaRL, and PODNet require relatively few parameters, their recognition accuracies are lower than those of the expansion-based methods. DS-AL stores no old-class exemplars but introduces a large parameter footprint through its analytic mappings. TagFex achieves strong accuracy but still retains a set of historical exemplars. In contrast, P-SER requires 11.99~M and 11.92~M parameters in the 4-stage and 2-stage settings, respectively, and stores no old-class samples while obtaining the highest final and average accuracies. These results confirm that the favorable balance among recognition performance, storage efficiency, and incremental stability observed in simulation is preserved under measured conditions.
\begin{table}[t]
	\caption{Performance Comparison on the Measured Dataset}
	\label{tab:comparison2}
	\centering
	\scriptsize
	\renewcommand{\arraystretch}{1.0}
	\resizebox{\columnwidth}{!}{
		\begin{tabular}{l@{\hspace{3pt}}c@{\hspace{3pt}}c@{\hspace{3pt}}c@{\hspace{3pt}}c@{\hspace{3pt}}c@{\hspace{3pt}}c@{\hspace{3pt}}c@{\hspace{3pt}}c}  
			\toprule
			\multirow{2}{*}{\textbf{Method}} 
			& \multirow{2}{*}{\shortstack{\textbf{Exemplars}\\\textbf{per class}}}
			& \multicolumn{3}{c}{\textbf{4-stage}} 
			& \multicolumn{3}{c}{\textbf{2-stage}} \\  
			\cmidrule(lr){3-5} \cmidrule(lr){6-8}  
			& 
			& \textbf{Params. (M)} & \textbf{$\mathcal{A}_{B}$ (\%)} & \textbf{$\bar{\mathcal{A}}$ (\%)} 
			& \textbf{Params. (M)} & \textbf{$\mathcal{A}_{B}$ (\%)} & \textbf{$\bar{\mathcal{A}}$ (\%)} \\  
			\midrule
			LwF\cite{Li2017} & 0 & 11.17 & 43.75 & 63.25 & 11.17 & 51.63 & 67.44 \\
			iCaRL\cite{Rebuffi2017} & 150 & 11.18 & 53.14 & 65.47 & 11.18 & 55.29 & 68.07 \\
			PODNet\cite{Douillard2020} & 150 & 11.18 & 48.62 & 63.81 & 11.18 & 52.81 & 64.90 \\
			DER\cite{Yan2021} & 150 & 44.71 & 60.47 & 69.32 & 22.37 & 65.47 & 72.12 \\
			VAE+DER\cite{Du2019,Yan2021} & 150 & 44.74 & 61.72 & 71.34 & 22.40 & 64.38 & 74.77 \\
			TAGNet+DER\cite{Liu2025,Yan2021} & 150 & 44.72 & 69.26 & 78.36 & 22.38 & 72.13 & 78.44 \\
			FOSTER\cite{Wang2022} & 150 & 35.55 & 59.38 & 72.70 & 17.78 & 68.76 & 76.29 \\
			DS-AL\cite{Zhuang2024} & 0 & 87.60 & 75.84 & 78.91 & 87.60 & 77.58 & 80.57 \\
			TagFex\cite{Zheng2025} & 100 & 35.94 & 74.31 & 80.69 & 18.71 & 75.34 & 80.45 \\
			P-SER & 0 & 11.99 & \textbf{76.93} & \textbf{82.30} & 11.92 & \textbf{79.85} & \textbf{83.56} \\
			\bottomrule
		\end{tabular}
	}
\end{table}
\begin{table*}[t]
	\centering
	\caption{Ablation study of the proposed components.}
	\label{tab:ablation}
	\scriptsize
	\setlength{\tabcolsep}{3.2pt}
	\renewcommand{\arraystretch}{1.15}
	\begin{tabular}{lccccccccccccc}
		\toprule
		\multirow{2}{*}{Setting} 
		& \multicolumn{3}{c}{Component}
		& \multicolumn{9}{c}{Recognition accuracy (\%)} 
		& \multirow{2}{*}{Avg.} \\
		\cmidrule(lr){2-4} \cmidrule(lr){5-13}
		& PCSP & PSD-SCIC & LRAE
		& Base & Stage-1 & Stage-2 & Stage-3 & Stage-4 & Stage-5 & Stage-6 & Stage-7 & Stage-8
		&  \\
		\midrule
		Ablation Exp-1 
		& $\times$ & $\times$ & $\times$
		& 83.44 & 67.36 & 63.84 & 61.11 & 57.47 & 55.82 & 49.34 & 47.93 & 46.59
		& 59.21 \\
		
		Ablation Exp-2 
		& $\checkmark$ & $\times$ & $\times$
		& 96.29 & 89.37 & 87.24 & 84.68 & 80.15 & 77.92 & 73.46 & 67.81 & 65.29
		& 80.25 \\
		
		Ablation Exp-3 
		& $\checkmark$ & $\checkmark$ & $\times$
		& 96.29 & 89.37 & 85.76 & 83.18 & 82.64 & 78.31 & 75.87 & 74.22 & 72.59
		& 82.03 \\
		
		Ablation Exp-4 
		& $\checkmark$ & $\times$ & $\checkmark$
		& 96.29 & 93.72 & 87.35 & 85.72 & 84.19 & 81.63 & 77.84 & 75.27 & 74.51
		& 84.06 \\
		
		Ablation Exp-5 
		& $\times$ & $\checkmark$ & $\times$
		& 83.44 & 67.36 & 73.28 & 69.91 & 65.43 & 59.75 & 55.12 & 53.86 & 50.34
		& 64.28 \\
		
		Ablation Exp-6 
		& $\times$ & $\checkmark$ & $\checkmark$
		& 83.44 & 82.39 & 85.21 & 83.69 & 82.44 & 80.18 & 76.73 & 73.95 & 72.10
		& 80.01 \\
		
		Ablation Exp-7 
		& $\times$ & $\times$ & $\checkmark$
		& 83.44 & 82.39 & 79.64 & 75.17 & 74.82 & 71.45 & 68.23 & 64.78 & 63.56
		& 73.72 \\
		\midrule
		Full P-SER
		& $\checkmark$ & $\checkmark$ & $\checkmark$
		& 96.29 & 93.72 & 91.97 & 88.94 & 87.62 & 85.41 & 84.15 & 82.96 & 81.35
		& \textbf{88.05} \\
		\bottomrule
	\end{tabular}
\end{table*}
\subsection{Ablation Studies}
To analyze the contribution of each component, PCSP, PSD-SCIC, and LRAE are separated and evaluated under the 8-stage protocol on the self-constructed simulation dataset. Seven controlled ablation settings are constructed from different module combinations, while the complete P-SER configuration is included as a reference. For controlled comparison, variants with the same trainable components share the corresponding model checkpoint, and only the module under investigation is switched on or off. The results are reported in \Cref{tab:ablation}.

The comparison between Ablation Exp-1 and Ablation Exp-2 demonstrates the independent contribution of PCSP to radar target recognition. Without PSD-SCIC and LRAE, introducing PCSP increases the average accuracy from 59.21\% to 80.25\%, corresponding to an absolute gain of 21.04\%. The stage-wise accuracy trajectory of Ablation Exp-2 is also substantially smoother than that of Ablation Exp-1. Although PCSP is not designed solely for the incremental process, its class-level physical scattering prototypes retain stable target properties for historical categories. These physical anchors reduce representation drift and consequently alleviate catastrophic forgetting to some extent.

The comparison between Ablation Exp-1 and Ablation Exp-5 reflects the inference calibration effect of PSD-SCIC. Because the first reflector class is introduced at Stage-2, PSD-SCIC is not activated at the Base and Stage-1 stages. The two settings therefore share the same checkpoint and produce identical results before Stage-2. From Stage-2 onward, the mechanism-level correction improves ship--reflector discrimination and raises the average accuracy from 59.21\% to 64.28\%, an absolute gain of 5.07\%. The same contribution is observed when PSD-SCIC is added to PCSP+LRAE: the average accuracy increases from 84.06\% in Ablation Exp-4 to 88.05\% in the complete model. These paired results isolate the calibration effect from changes in feature training. Ablation Exp-7 independently verifies LRAE. Introducing only lightweight adapter expansion increases the average accuracy from 59.21\% to 73.72\%, showing that stage-specific residual capacity improves new-class adaptation without duplicating the complete backbone.

Ablation Exp-3 combines PCSP and PSD-SCIC and reaches an average accuracy of 82.03\% by jointly addressing aspect sensitivity and reflector-induced ambiguity. Nevertheless, because it lacks an incremental expansion mechanism, its accuracy still decreases markedly as the number of stages grows. Ablation Exp-4 and Ablation Exp-6 include LRAE and therefore maintain more stable recognition during later updates. Among the two-component configurations, PCSP+LRAE in Ablation Exp-4 achieves the highest average accuracy of 84.06\%, confirming the complementarity between stable physical constraints and parameter-isolated representation expansion.

Finally, the complete P-SER model achieves an average accuracy of 88.05\%. The ablation results demonstrate that the three components address different limitations rather than providing redundant improvements: PCSP stabilizes aspect-sensitive radar representations, PSD-SCIC calibrates ship--reflector ambiguity at inference, and LRAE controls forgetting and model growth through lightweight expansion. Their combination therefore improves both recognition accuracy and incremental stability.

\subsection{The Target Reconstruction Experiment and Model Analysis}

\begin{figure}[b]
	\centering
	\includegraphics[width=\columnwidth]{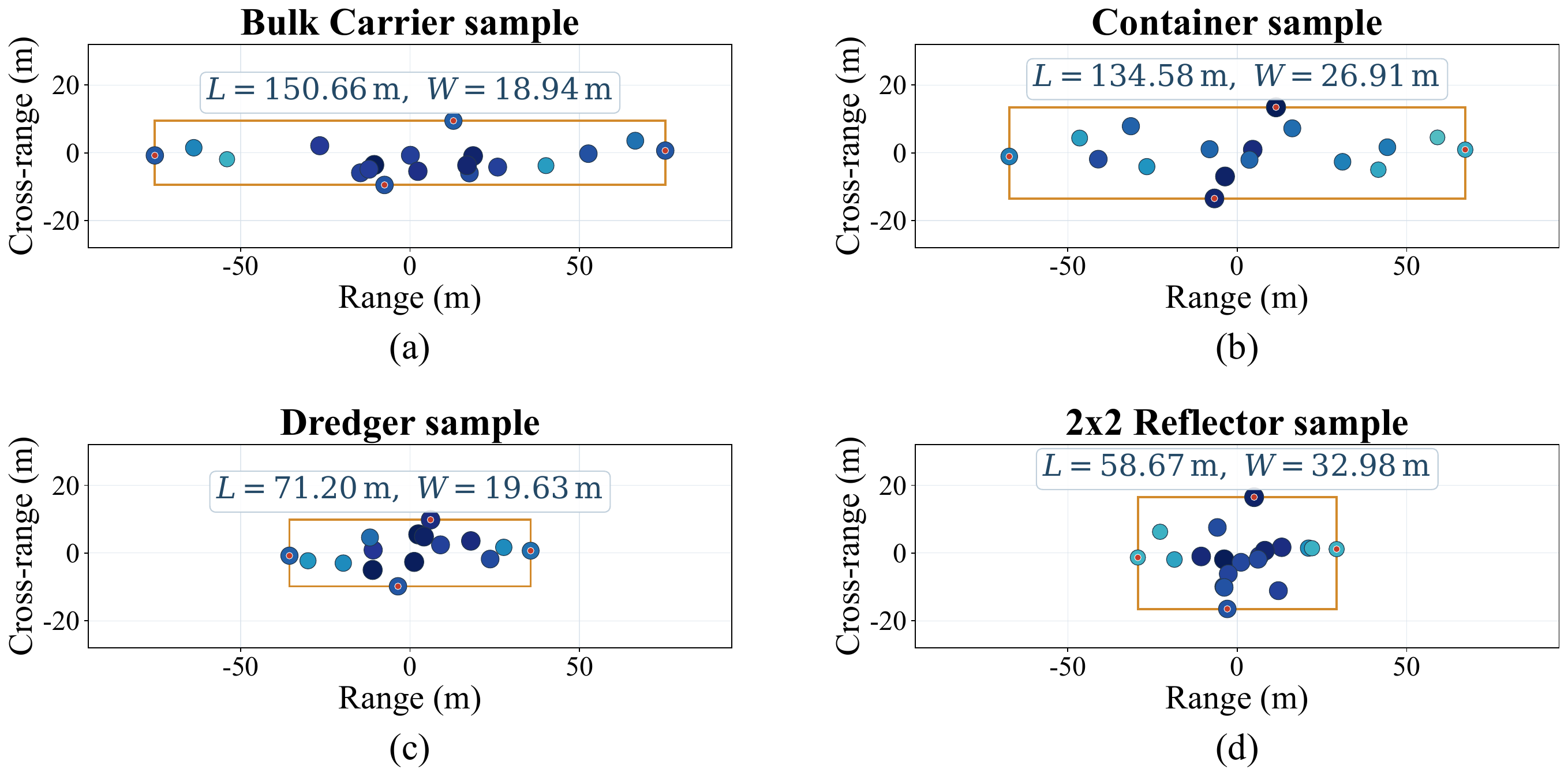}
	\caption{Target reconstruction results for representative observations from the simulation test set. The dots denote the retained dominant scattering points, and the orange rectangles indicate the reconstructed target extents in the range and cross-range directions.}
	\label{fig:target_reconstruction}
\end{figure}

To examine the intermediate physical representation produced by PCSP, \Cref{fig:target_reconstruction} visualizes four representative observations drawn from the simulation test set. One sample is selected from each target class shown, so the figure presents sample-level examples rather than complete-test-set statistics. The dots denote dominant scattering points retained after consistency filtering, and the bounding rectangles indicate the estimated target extents.

The bulk-carrier and container examples recover the principal target extents relatively well because their dominant responses cover the bow, stern, sides, and deck boundaries. The dredger shows a larger deviation owing to its smaller size and nonuniform scattering distribution. The $2\times2$ reflector is more sensitive to aspect and peak selection because only a few specular and multiple-bounce components define its apparent boundary. These examples illustrate both the physical meaning and the expected limitations of sparse scattering-point reconstruction.

Beyond these examples, \Cref{tab:reconstruction_statistics} reports the classwise mean dimensions reconstructed from all samples in the simulation test set. The relative errors are computed separately as $\epsilon_L=|\widehat{L}-L|/L\times100\%$ and $\epsilon_W=|\widehat{W}-W|/W\times100\%$, using the dimensions in \Cref{tab:datasets} as ground truth.
\begin{table}[t]
	\caption{Full-Test-Set Reconstruction Statistics on the Simulation Dataset}
	\label{tab:reconstruction_statistics}
	\centering
	\scriptsize
	\setlength{\tabcolsep}{3pt}
	\renewcommand{\arraystretch}{1.05}
	\resizebox{\columnwidth}{!}{
	\begin{tabular}{lcccc}
		\toprule
		\textbf{Target} & \textbf{GT $(L,W)$ (m)} & \textbf{Reconstructed $(\widehat{L},\widehat{W})$ (m)} & \textbf{$\epsilon_L$ (\%)} & \textbf{$\epsilon_W$ (\%)} \\
		\midrule
		Bulk carrier & (152.14, 16.07) & (149.38, 18.62) & 1.81 & 15.87 \\
		Container & (139.37, 23.37) & (141.66, 25.97) & 1.64 & 11.13 \\
		Passenger & (178.82, 26.01) & (186.59, 29.50) & 4.35 & 13.42 \\
		General cargo & (121.50, 21.72) & (117.67, 23.09) & 3.15 & 6.31 \\
		Dredger & (79.52, 15.42) & (83.25, 17.84) & 4.69 & 15.69 \\
		Fishing & (46.85, 8.60) & (50.13, 11.81) & 7.00 & 37.33 \\
		Tanker & (168.55, 18.44) & (184.26, 20.38) & 9.32 & 10.52 \\
		$1\times2$ reflector & (44.40, 7.20) & (63.04, 11.45) & 41.98 & 59.03 \\
		$2\times2$ reflector & (44.40, 24.40) & (68.85, 21.66) & 55.07 & 11.23 \\
		$2\times3$ reflector & (81.60, 24.40) & (83.21, 30.23) & 1.97 & 23.89 \\
		\midrule
		Class mean & -- & -- & 13.10 & 20.44 \\
		\bottomrule
	\end{tabular}}
\end{table}

For the seven ship classes, the mean length and width errors are 4.57\% and 15.75\%, respectively. The lower length error indicates that stable bow--stern responses provide a reliable estimate of longitudinal extent, whereas width estimation depends more strongly on monopulse angle accuracy and the visibility of cross-range boundaries. The fishing vessel consequently has a larger width error because its small physical width is represented by fewer dominant components. The three reflector configurations yield higher mean errors of 33.01\% and 31.38\%. Their sparse, aspect-dependent responses do not necessarily cover the geometric boundary, although the $2\times3$ reflector still attains a length error of only 1.97\%. These results support the use of reconstructed attributes as aggregated physical priors rather than precision geometric measurements.

\begin{figure}[t]
	\centering
	\includegraphics[width=\columnwidth]{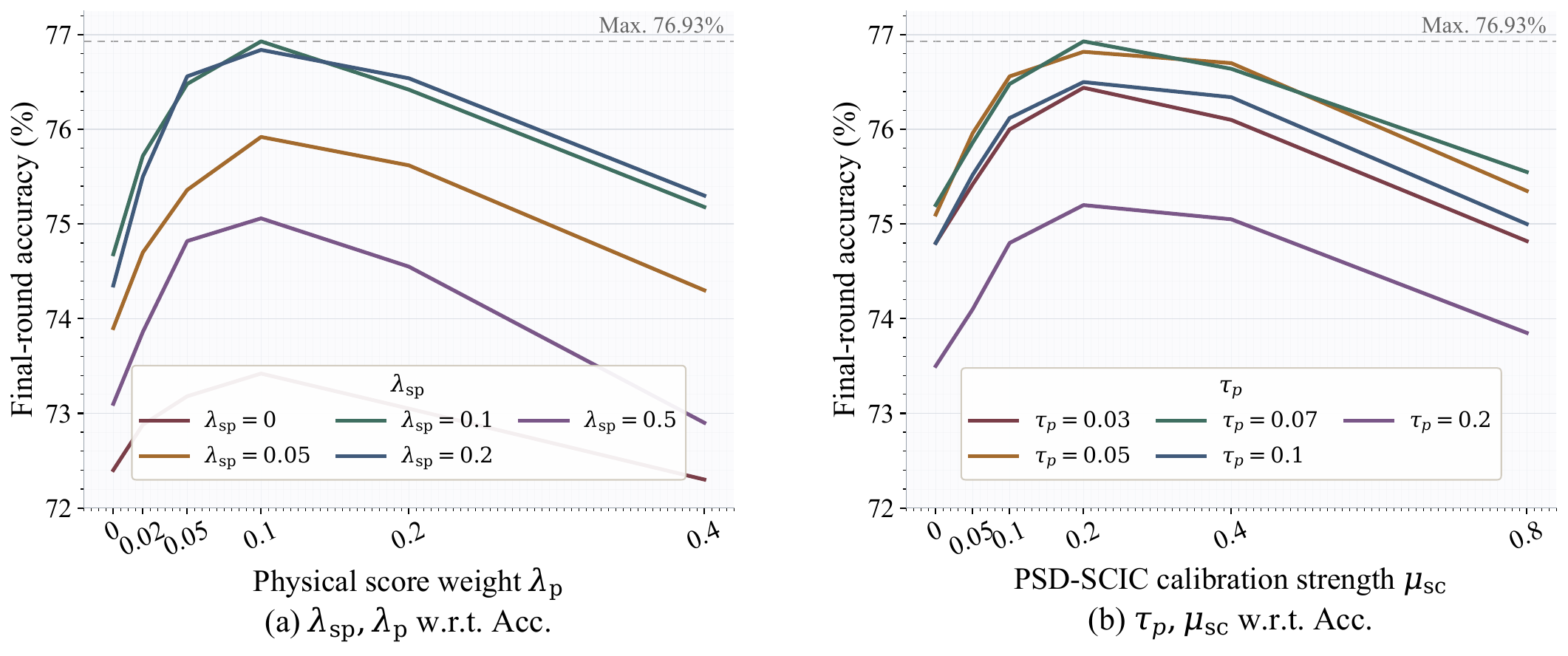}
	\caption{Hyperparameter sensitivity on the measured 4-stage protocol: final-round accuracy with respect to (a) $\lambda_{\mathrm{sp}}$ and $\lambda_{\mathrm{p}}$, and (b) $\tau_p$ and $\mu_{\mathrm{sc}}$.}
	\label{fig:measured_hyperparameter_analysis}
\end{figure}

The sensitivity of four central hyperparameters is evaluated on the measured dataset in \Cref{fig:measured_hyperparameter_analysis}. The dashed line denotes the best final-round accuracy of 76.93\%. In \Cref{fig:measured_hyperparameter_analysis}(a), moderate $\lambda_{\mathrm{sp}}$ and $\lambda_{\mathrm{p}}$ values improve recognition by incorporating prototype guidance, whereas overly large values overemphasize the reconstructed prior. In \Cref{fig:measured_hyperparameter_analysis}(b), $\tau_p$ controls the sharpness of prototype matching and $\mu_{\mathrm{sc}}$ determines the PSD calibration strength. Moderate settings improve ambiguous ship--reflector decisions, while excessive smoothing or correction weakens the original discriminative evidence. Overall, the results favor a balanced contribution from the physical prototypes and inference calibration.

\section{Conclusion}

This paper presented P-SER for class-incremental radar target recognition. PCSP reconstructs stable scattering attributes to constrain representation learning, PSD-SCIC checks ambiguous ship--reflector predictions through scattering mechanisms, and LRAE expands the model with compact weighted residual adapters. Experiments on simulated and measured datasets show that P-SER improves incremental accuracy and stability without retaining old-class exemplars, while reconstruction, confusion-matrix, ablation, and sensitivity analyses validate its individual components.

Future work will enrich PCSP with scattering-component--semantic-description pairs to improve interpretability and transferability. PSD-SCIC will also be extended from ship--reflector ambiguity to all target classes through uncertainty-aware mechanism calibration.

\bibliographystyle{IEEEtaes}
\Needspace{6\baselineskip}
\bibliography{hrrp2}

@Article{Kirkpatrick2017,
  author    = {Kirkpatrick, James and Pascanu, Razvan and Rabinowitz, Neil and Veness, Joel and Desjardins, Guillaume and Rusu, Andrei A and Milan, Kieran and Quan, John and Ramalho, Tiago and Grabska-Barwinska, Agnieszka and others},
  journal   = {Proc. Natl. Acad. Sci.},
  title     = {Overcoming catastrophic forgetting in neural networks},
  year      = {2017},
  number    = {13},
  pages     = {3521--3526},
  volume    = {114},
  publisher = {National Academy of Sciences},
}

@InProceedings{Rebuffi2017,
  author    = {Rebuffi, S. and Kolesnikov, Alexander and Sperl, Georg and Lampert, Christoph H},
  booktitle = {Proc. IEEE Conf. Comput. Vis. Pattern Recognit. (CVPR)},
  title     = {{iCaRL}: Incremental classifier and representation learning},
  year      = {2017},
  pages     = {2001--2010},
}

@InProceedings{Bang2022,
  author    = {Bang, Jihwan and Koh, Hyunseo and Park, Seulki and Song, Hwanjun and Ha, Jung-Woo and Choi, Jonghyun},
  booktitle = {Proc. IEEE Conf. Comput. Vis. Pattern Recognit. (CVPR)},
  title     = {Online continual learning on a contaminated data stream with blurry task boundaries},
  year      = {2022},
  pages     = {9275--9284},
}

@InProceedings{DeLange2021,
  author    = {De Lange, Matthias and Tuytelaars, Tinne},
  booktitle = {Proc. IEEE Int. Conf. Comput. Vis. (ICCV)},
  title     = {Continual prototype evolution: Learning online from non-stationary data streams},
  year      = {2021},
  pages     = {8250--8259},
}

@Article{Hinton2015,
  author  = {Hinton, Geoffrey and Vinyals, Oriol and Dean, Jeff},
  journal = {arXiv preprint arXiv:1503.02531},
  title   = {Distilling the knowledge in a neural network},
  year    = {2015},
}

@Article{Li2017,
  author    = {Li, Zhizhong and Hoiem, Derek},
  journal   = {IEEE Trans. Pattern Anal. Mach. Intell.},
  title     = {Learning without forgetting},
  year      = {2017},
  number    = {12},
  pages     = {2935--2947},
  volume    = {40},
  publisher = {IEEE},
}

@InProceedings{Douillard2020,
  author       = {Douillard, Arthur and Cord, Matthieu and Ollion, Charles and Robert, Thomas and Valle, Eduardo},
  booktitle    = {Proc. Eur. Conf. Comput. Vis. (ECCV)},
  title        = {{PODNet}: Pooled outputs distillation for small-tasks incremental learning},
  year         = {2020},
  organization = {Springer},
  pages        = {86--102},
}

@InProceedings{Kang2022,
  author    = {Kang, Minsoo and Park, Jaeyoo and Han, Bohyung},
  booktitle = {Proc. IEEE Conf. Comput. Vis. Pattern Recognit. (CVPR)},
  title     = {Class-incremental learning by knowledge distillation with adaptive feature consolidation},
  year      = {2022},
  pages     = {16071--16080},
}

@Article{Lee2017,
  author  = {Lee, Jeongtae and Yoon, Jaehong and Yang, Eunho and Hwang, Sung Ju},
  journal = {CoRR},
  title   = {Lifelong learning with dynamically expandable networks},
  year    = {2017},
}

@InProceedings{Yan2021,
  author    = {Yan, Shipeng and Xie, Jiangwei and He, Xuming},
  booktitle = {Proc. IEEE Conf. Comput. Vis. Pattern Recognit. (CVPR)},
  title     = {{DER}: Dynamically expandable representation for class-incremental learning},
  year      = {2021},
  pages     = {3014--3023},
}

@InProceedings{Wang2022,
  author       = {Wang, F. and Zhou, D. and Ye, H. and Zhan, D.},
  booktitle    = {Proc. Eur. Conf. Comput. Vis. (ECCV)},
  title        = {{FOSTER}: Feature boosting and compression for class-incremental learning},
  year         = {2022},
  organization = {Springer},
  pages        = {398--414},
}

@Article{Tai2023,
  author    = {Tai, Yuan and Tan, Yihua and Xiong, Shengzhou and Tian, Jinwen},
  journal   = {IEEE Trans. Geosci. Remote Sens.},
  title     = {Mine-distill-prototypes for complete few-shot class-incremental learning in image classification},
  year      = {2023},
  pages     = {1--13},
  volume    = {61},
  publisher = {IEEE},
}

@Article{Chen2020,
  author    = {Chen, Jingzhou and Wang, Shihao and Chen, Ling and Cai, Haibin and Qian, Yuntao},
  journal   = {IEEE Trans. Geosci. Remote Sens.},
  title     = {Incremental detection of remote sensing objects with feature pyramid and knowledge distillation},
  year      = {2020},
  pages     = {1--13},
  volume    = {60},
  publisher = {IEEE},
}

@Article{Pilcher2011,
  author    = {Pilcher, Christopher M and Khotanzad, Alireza},
  journal   = {IEEE Trans. Aerosp. Electron. Syst.},
  title     = {Maritime {ATR} using classifier combination and high-resolution range profiles},
  year      = {2011},
  number    = {4},
  pages     = {2558--2573},
  volume    = {47},
  publisher = {IEEE},
}

@Article{Du2015,
  author    = {Du, Lan and Wang, Penghui and Zhang, Lei and He, Hua and Liu, Hongwei},
  journal   = {Expert Syst. Appl.},
  title     = {Robust statistical recognition and reconstruction scheme based on hierarchical {Bayesian} learning of {HRR} radar target signal},
  year      = {2015},
  number    = {14},
  pages     = {5860--5873},
  volume    = {42},
  publisher = {Elsevier},
}

@Article{Zhu2007,
  author    = {Zhu, Feng and Zhang, Xian-Da and Hu, Ya-Feng and Xie, Deguang},
  journal   = {IEEE Trans. Signal Process.},
  title     = {Nonstationary hidden {Markov} models for multiaspect discriminative feature extraction from radar targets},
  year      = {2007},
  number    = {5},
  pages     = {2203--2214},
  volume    = {55},
  publisher = {IEEE},
}

@Article{Du2008,
  author    = {Du, Lan and Liu, Hongwei and Bao, Zheng},
  journal   = {IEEE Trans. Signal Process.},
  title     = {Radar {HRRP} statistical recognition: Parametric model and model selection},
  year      = {2008},
  number    = {5},
  pages     = {1931--1944},
  volume    = {56},
  publisher = {IEEE},
}

@Article{Kim2002,
  author    = {Kim, Kyung-Tae and Seo, Dong-Kyu and Kim, Hyo-Tae},
  journal   = {IEEE Trans. Antennas Propag.},
  title     = {Efficient radar target recognition using the {MUSIC} algorithm and invariant features},
  year      = {2002},
  number    = {3},
  pages     = {325--337},
  volume    = {50},
  publisher = {IEEE},
}

@InProceedings{Slomka1999,
  author       = {Slomka, Stefan and Gibbins, Danny and Gray, D and Haywood, Brett},
  booktitle    = {Proc. 5th Int. Symp. Signal Process. Appl. (ISSPA)},
  title        = {Features for high-resolution radar range-profile-based ship classification},
  year         = {1999},
  pages        = {329--332},
  volume       = {1},
}

@Article{Wan2019,
  author    = {Wan, Jinwei and Chen, Bo and Xu, Bin and Liu, Hongwei and Jin, Lin},
  journal   = {EURASIP J. Adv. Signal Process.},
  title     = {Convolutional neural networks for radar {HRRP} target recognition and rejection},
  year      = {2019},
  number    = {1},
  pages     = {5},
  volume    = {2019},
  publisher = {Springer},
}

@Article{Wan2020,
  author    = {Wan, Jinwei and Chen, Bo and Liu, Yingqi and Yuan, Yijun and Liu, Hongwei and Jin, Lin},
  journal   = {IEEE Access},
  title     = {Recognizing the {HRRP} by combining {CNN} and {BiRNN} with attention mechanism},
  year      = {2020},
  pages     = {20828--20837},
  volume    = {8},
  publisher = {IEEE},
}

@Article{Liao2018,
  author    = {Liao, Kuo and Si, Jinxiu and Zhu, Fangqi and He, Xudong},
  journal   = {IEEE Access},
  title     = {Radar {HRRP} target recognition based on concatenated deep neural networks},
  year      = {2018},
  pages     = {29211--29218},
  volume    = {6},
  publisher = {IEEE},
}

@Article{Guo2019,
  author    = {Guo, Chen and He, You and Wang, Haipeng and Jian, Tao and Sun, Shun},
  journal   = {IEEE Access},
  title     = {Radar {HRRP} target recognition based on deep one-dimensional residual-inception network},
  year      = {2019},
  pages     = {9191--9204},
  volume    = {7},
  publisher = {IEEE},
}

@Article{Su2023,
  author    = {Su, Keyu and Gong, Lin and Wang, Guanyong and Zhang, Lei},
  journal   = {IEEE Geosci. Remote Sens. Lett.},
  title     = {Noise-robust radar {HRRP} target sequential recognition based on correlative scattering centers},
  year      = {2023},
  pages     = {1--5},
  volume    = {20},
  publisher = {IEEE},
}

@Article{Zeng2022,
  author    = {Zeng, Zhiqiang and Sun, Jinping and Han, Zhu and Hong, Wen},
  journal   = {IEEE Geosci. Remote Sens. Lett.},
  title     = {Radar {HRRP} target recognition method based on multi-input convolutional gated recurrent unit with cascaded feature fusion},
  year      = {2022},
  pages     = {1--5},
  volume    = {19},
  publisher = {IEEE},
}

@Article{Chen2021,
  author    = {Chen, Wenchao and Chen, Bo and Peng, Xiaojun and Liu, Jiaqi and Yang, Yang and Zhang, Hao and Liu, Hongwei},
  journal   = {IEEE Trans. Signal Process.},
  title     = {Tensor {RNN} with {Bayesian} nonparametric mixture for radar {HRRP} modeling and target recognition},
  year      = {2021},
  pages     = {1995--2009},
  volume    = {69},
  publisher = {IEEE},
}

@Article{Guo2020a,
  author    = {Guo, Dandan and Chen, Bo and Chen, Wenchao and Wang, Chaojie and Liu, Hongwei and Zhou, Mingyuan},
  journal   = {IEEE Trans. Signal Process.},
  title     = {Variational temporal deep generative model for radar {HRRP} target recognition},
  year      = {2020},
  pages     = {5795--5809},
  volume    = {68},
  publisher = {IEEE},
}

@Article{Chen2021a,
  author    = {Chen, Wenchao and Chen, Bo and Liu, Yicheng and Peng, Xiaojun and Fan, Haoyang and Yu, Fangxu and Liu, Hongwei},
  journal   = {Signal Process.},
  title     = {Bidirectional recurrent gamma belief network for {HRRP} target recognition},
  year      = {2021},
  pages     = {108213},
  volume    = {188},
  publisher = {Elsevier},
}

@Article{Zhang2020,
  author    = {Zhang, Yifan and Xiao, Fei and Qian, Fengchen and Li, Xiang},
  journal   = {IEEE Access},
  title     = {{VGM-RNN}: {HRRP} sequence extrapolation and recognition based on a novel optimized {RNN}},
  year      = {2020},
  pages     = {70071--70081},
  volume    = {8},
  publisher = {IEEE},
}

@Article{Kingma2013,
  author  = {Kingma, Diederik P and Welling, Max},
  journal = {arXiv preprint arXiv:1312.6114},
  title   = {Auto-encoding variational {Bayes}},
  year    = {2013},
}

@Article{Du2019,
  author    = {Du, Chuan and Chen, Bo and Xu, Bin and Guo, Dandan and Liu, Hongwei},
  journal   = {Signal Process.},
  title     = {Factorized discriminative conditional variational auto-encoder for radar {HRRP} target recognition},
  year      = {2019},
  pages     = {176--189},
  volume    = {158},
  publisher = {Elsevier},
}

@Article{Liao2021,
  author    = {Liao, Leiyao and Du, Lan and Chen, Jian},
  journal   = {Signal Process.},
  title     = {Class factorized complex variational auto-encoder for {HRR} radar target recognition},
  year      = {2021},
  pages     = {107932},
  volume    = {182},
  publisher = {Elsevier},
}

@Article{Chen2020a,
  author    = {Chen, Jian and Du, Lan and Liao, Leiyao},
  journal   = {IEEE Trans. Cybern.},
  title     = {Discriminative mixture variational autoencoder for semisupervised classification},
  year      = {2020},
  number    = {5},
  pages     = {3032--3046},
  volume    = {52},
  publisher = {IEEE},
}

@Article{Huang2021,
  author    = {Huang, Yue and Wen, Yi and Shi, Liangchao and Ding, Xinghao},
  journal   = {IEEE Geosci. Remote Sens. Lett.},
  title     = {Recognition-aware {HRRP} generation with generative adversarial network},
  year      = {2021},
  pages     = {1--5},
  volume    = {19},
  publisher = {IEEE},
}

@Article{Li2023,
  author    = {Li, Xungen and Ouyang, Wenqing and Pan, Mian and Lv, Shuaishuai and Ma, Qi},
  journal   = {IEEE Trans. Geosci. Remote Sens.},
  title     = {Continuous learning method of radar {HRRP} based on {CVAE-GAN}},
  year      = {2023},
  pages     = {1--19},
  volume    = {61},
  publisher = {IEEE},
}

@Article{Yin2024,
  author    = {Yin, Jianguo and Sheng, Wen and Jiang, He},
  journal   = {IEEE Access},
  title     = {Small-sample target recognition based on radar {HRRP} and {SDAE-WACGAN}},
  year      = {2024},
  pages     = {16375--16385},
  volume    = {12},
  publisher = {IEEE},
}

@InProceedings{Zhou2024,
  author       = {Zhou, Jiaqi and Li, Jixi and Guo, Weiwei and Zheng, Xin and Yu, Wenxian},
  booktitle    = {Proc. IEEE Int. Geosci. Remote Sens. Symp. (IGARSS)},
  title        = {Knowledge-guided {BiTCN} prototypical network for few-shot radar {HRRP} target recognition},
  year         = {2024},
  pages        = {10223--10227},
}

@Article{Li2025a,
  author    = {Li, Jixi and Guo, Weiwei and Zhou, Jiaqi and Wei, Feiming and Zhang, Tao and Yu, Wenxian},
  journal   = {IEEE Trans. Aerosp. Electron. Syst.},
  title     = {Patching-based {HRRP} nonexemplar class-incremental recognition via {Gaussian} truncation},
  year      = {2025},
  doi       = {10.1109/TAES.2025.3625527},
  publisher = {IEEE},
}

@Article{Liu2025,
  author    = {Liu, Qi and Zhang, Xinyu and Liu, Yongxiang},
  journal   = {IEEE Trans. Aerosp. Electron. Syst.},
  title     = {Target-aspect-guided neural network with geometric constraints for imbalanced radar target recognition},
  year      = {2025},
  number    = {5},
  pages     = {12122--12142},
  volume    = {61},
  doi       = {10.1109/TAES.2025.3573273},
  publisher = {IEEE},
}

@InProceedings{Zhuang2024,
  author    = {Zhuang, Huiping and He, Run and Tong, Kai and Zeng, Ziqian and Chen, Cen and Lin, Zhiping},
  booktitle = {Proc. AAAI Conf. Artif. Intell. (AAAI)},
  title     = {{DS-AL}: A dual-stream analytic learning method for exemplar-free class-incremental learning},
  year      = {2024},
  number    = {15},
  pages     = {17237--17244},
  volume    = {38},
}

@InProceedings{Petit2023,
  author    = {Petit, Gr{\'e}goire and Popescu, Adrian and Schindler, Hugo and Picard, David and Delezoide, Bertrand},
  booktitle = {Proc. IEEE Winter Conf. Appl. Comput. Vis. (WACV)},
  title     = {{FeTrIL}: Feature translation for exemplar-free class-incremental learning},
  year      = {2023},
  pages     = {3911--3920},
}

@Article{Li2025c,
  author    = {Li, Zilin and Fu, Jidai and Li, Wentao and Li, Shuai and Tian, Biao and Xu, Shiyou},
  journal   = {IEEE Sensors J.},
  title     = {{PFDE-DCN}: A lightweight class-incremental learning method for radar {HRRP} recognition},
  year      = {2025},
  doi       = {10.1109/JSEN.2025.3604333},
  publisher = {IEEE},
}

@InProceedings{Chang2025,
  author       = {Chang, Zhipeng and Lv, Yuzhou and Tian, Bo and Xiao, Xianjie and Guo, Shuai and Wang, Penghui},
  booktitle    = {Proc. 5th Int. Conf. Commun. Technol. Inf. Technol. (ICCTIT)},
  title        = {An Incremental Target Recognition Method for HRRP Based on Autoencoder Reconstruction and Multi-Level Knowledge Distillation},
  year         = {2025},
  pages        = {363--369},
}

@InProceedings{Zheng2025,
  author    = {Zheng, Bowen and Zhou, D. and Ye, H. and Zhan, D.},
  booktitle = {Proc. IEEE Conf. Comput. Vis. Pattern Recognit. (CVPR)},
  title     = {Task-agnostic guided feature expansion for class-incremental learning},
  year      = {2025},
  pages     = {10099--10109},
}

@Article{Zhang2023,
  author    = {Zhang, Xu and Xu, Feng and Yang, Ying and Xing, Mengdao and Jin, Yaqiu},
  journal   = {IEEE Trans. Antennas Propag.},
  title     = {A primitive scatterer dictionary for semantic representation of radar target images},
  year      = {2024},
  number    = {1},
  pages     = {825--840},
  volume    = {72},
  doi       = {10.1109/TAP.2023.3321386},
  publisher = {IEEE},
}

@Article{Chen2025,
  author    = {Chen, Junyan and Li, Wentao and Li, Shuai and Tian, Biao and Chen, Zengping},
  journal   = {IEEE Trans. Aerosp. Electron. Syst.},
  title     = {{ECAPL}: Open set recognition on {HRRP} through expansive consistency-aware prototypes},
  year      = {2025},
  publisher = {IEEE},
}

@Article{Yang2025,
  author    = {Yang, Jieyuan and Cai, Zheyuan and Wang, Jiaxiang and He, Xuxiang and Zhong, Yijin and Cheng, Qiang and Sun, Jinming and Yang, Yuhao and Huang, Yue},
  journal   = {IEEE Trans. Aerosp. Electron. Syst.},
  title     = {A contrastive uncertainty modeling for radar {HRRP} domain generalization target recognition},
  year      = {2025},
  publisher = {IEEE},
}

@Article{Li2025d,
  author    = {Li, Wentao and Tian, Biao and Ma, Jing and Huang, Pengjun and Xu, Shiyou},
  journal   = {IEEE Trans. Aerosp. Electron. Syst.},
  title     = {{OSFSM}: A systematic open set framework for radar automatic target recognition using {HRRP}},
  year      = {2025},
  publisher = {IEEE},
}

@Article{Zhang2025,
  author    = {Zhang, Yanjin and Feng, Xuejian and Yin, Hongcheng and Wei, Xiao and Yan, Hua},
  journal   = {IEEE Trans. Aerosp. Electron. Syst.},
  title     = {{HRRPSeqNet}: Open-set recognition of space target motions using {HRRP} sequences},
  year      = {2025},
  publisher = {IEEE},
}

@Article{Xia2023,
  author    = {Xia, Ziheng and Wang, Penghui and Dong, Ganggang and Liu, Hongwei},
  journal   = {IEEE Trans. Geosci. Remote Sens.},
  title     = {Radar {HRRP} open set recognition based on extreme value distribution},
  year      = {2023},
  pages     = {1--16},
  volume    = {61},
  publisher = {IEEE},
}

@Article{Zhang2023a,
  author    = {Zhang, Yuan-Peng and Zhang, Lei and Kang, Le and Wang, Huan and Luo, Ying and Zhang, Qun},
  journal   = {IEEE Trans. Geosci. Remote Sens.},
  title     = {Space target classification with corrupted {HRRP} sequences based on temporal--spatial feature aggregation network},
  year      = {2023},
  pages     = {1--18},
  volume    = {61},
  publisher = {IEEE},
}

@Article{Li2025e,
  author    = {Li, Hongxu and Li, Xiaodi and Xu, Zihan and Jin, Xinfei and Gao, Jianjun and Su, Fulin},
  journal   = {IEEE Trans. Geosci. Remote Sens.},
  title     = {An end-to-end multi-domain interaction deep unrolling network based on block-aware optimization model for {ISAR} multi-target separation},
  year      = {2025},
  publisher = {IEEE},
}

@Article{Zhang2026,
  author    = {Zhang, Fan and Xu, Yuelei and Li, Huafeng and Luo, Wei and Zhu, Chunjia and Xia, Sijia and Zhang, Zhaoxiang},
  journal   = {IEEE Trans. Aerosp. Electron. Syst.},
  title     = {Causal-inspired unsupervised dynamic-attribute feature disentanglement network for {HRRP} target recognition},
  year      = {2026},
  publisher = {IEEE},
}

@Article{Wang2025,
  author    = {Wang, Yanhua and Liu, Yichen and Yu, Binhua and Wang, Weijia and Zhang, Liang and Zhang, Xin and Li, Yang},
  journal   = {IEEE Trans. Aerosp. Electron. Syst.},
  title     = {{NHPL}: Neural collapse guided hierarchical prototypical learning for radar {HRRP} open set recognition},
  year      = {2025},
  publisher = {IEEE},
}

@Article{Liu2025a,
  author    = {Liu, Mei and Gao, Xunzhang and Qiu, Xiangfeng and Liu, Yun},
  journal   = {IEEE Trans. Aerosp. Electron. Syst.},
  title     = {Deep {Gaussian} hidden {Markov} network for robust {HRRP} sequence modeling and target recognition},
  year      = {2025},
  publisher = {IEEE},
}

@Article{Yang2026,
  author    = {Yang, Wei and Wang, Weijia and Wang, Yanhua and Zhang, Liang and Zhang, Xin and Han, Yuqi and Li, Yang},
  journal   = {IEEE Trans. Aerosp. Electron. Syst.},
  title     = {Cross-frequency feature learning for dual-band radar {HRRP} target recognition},
  year      = {2026},
  publisher = {IEEE},
}

@Article{Wang2026,
  author    = {Wang, Ke and Pan, Xiaoyi and Ai, Xiaofeng and Xu, Zhiming and Zhao, Feng and Xiao, Shunping and Chen, Siwei and Song, Yafei},
  journal   = {IEEE Trans. Aerosp. Electron. Syst.},
  title     = {{Adv-STMamba}: A lightweight spatial--temporal {Mamba} model incorporating adversarial training for radar {HRRP} target recognition},
  year      = {2026},
  publisher = {IEEE},
}

@Article{Han2026,
  author    = {Han, Tao and Chen, Hui and Jia, Yizhen and Zheng, Zhi and Wang, Wen-Qin},
  journal   = {IEEE Trans. Aerosp. Electron. Syst.},
  title     = {Weighted exponential-product fusion and recursive {Bayesian} update for multi-radar cooperative target recognition},
  year      = {2026},
  publisher = {IEEE},
}

@Article{Li2026,
  author    = {Li, Wentao and Chen, Junyan and Li, Shuai and Tian, Biao and Chen, Zengping},
  journal   = {IEEE Trans. Aerosp. Electron. Syst.},
  title     = {A synthetic framework for real-world radar target recognition using {HRRP}},
  year      = {2026},
  publisher = {IEEE},
}

@Article{Li2026a,
  author    = {Li, Shuai and Li, Wentao and Chen, Junyan and Tian, Biao and Xu, Shiyou},
  journal   = {IEEE Trans. Instrum. Meas.},
  title     = {Structure-aware generative modeling and comprehensive evaluation for high-fidelity radar {HRRP} synthesis},
  year      = {2026},
  publisher = {IEEE},
}

@Article{Sun2026,
  author    = {Sun, Guozheng and Wang, Lei and Wang, Yanhao and Wang, Jie and Liu, Yimin},
  journal   = {IEEE Trans. Aerosp. Electron. Syst.},
  title     = {Point-spread-function guided {HRRP} recognition network: A framework for robust target classification under {ISRJ}},
  year      = {2026},
  publisher = {IEEE},
}

@Book{Sherman2011,
  author    = {Sherman, Samuel M. and Barton, David K.},
  publisher = {Artech House},
  title     = {Monopulse Principles and Techniques},
  year      = {2011},
  edition   = {2nd},
}

@Article{Nickel2006,
  author  = {Nickel, Ulrich},
  journal = {IEEE Aerosp. Electron. Syst. Mag.},
  title   = {Overview of generalized monopulse estimation},
  year    = {2006},
  volume  = {21},
  number  = {6},
  pages   = {27--56},
  doi     = {10.1109/MAES.2006.1662039},
}
\begin{IEEEbiography}[{\includegraphics[width=1in, height=1.1in,
		trim=0 10pt 0 12pt, clip, keepaspectratio]{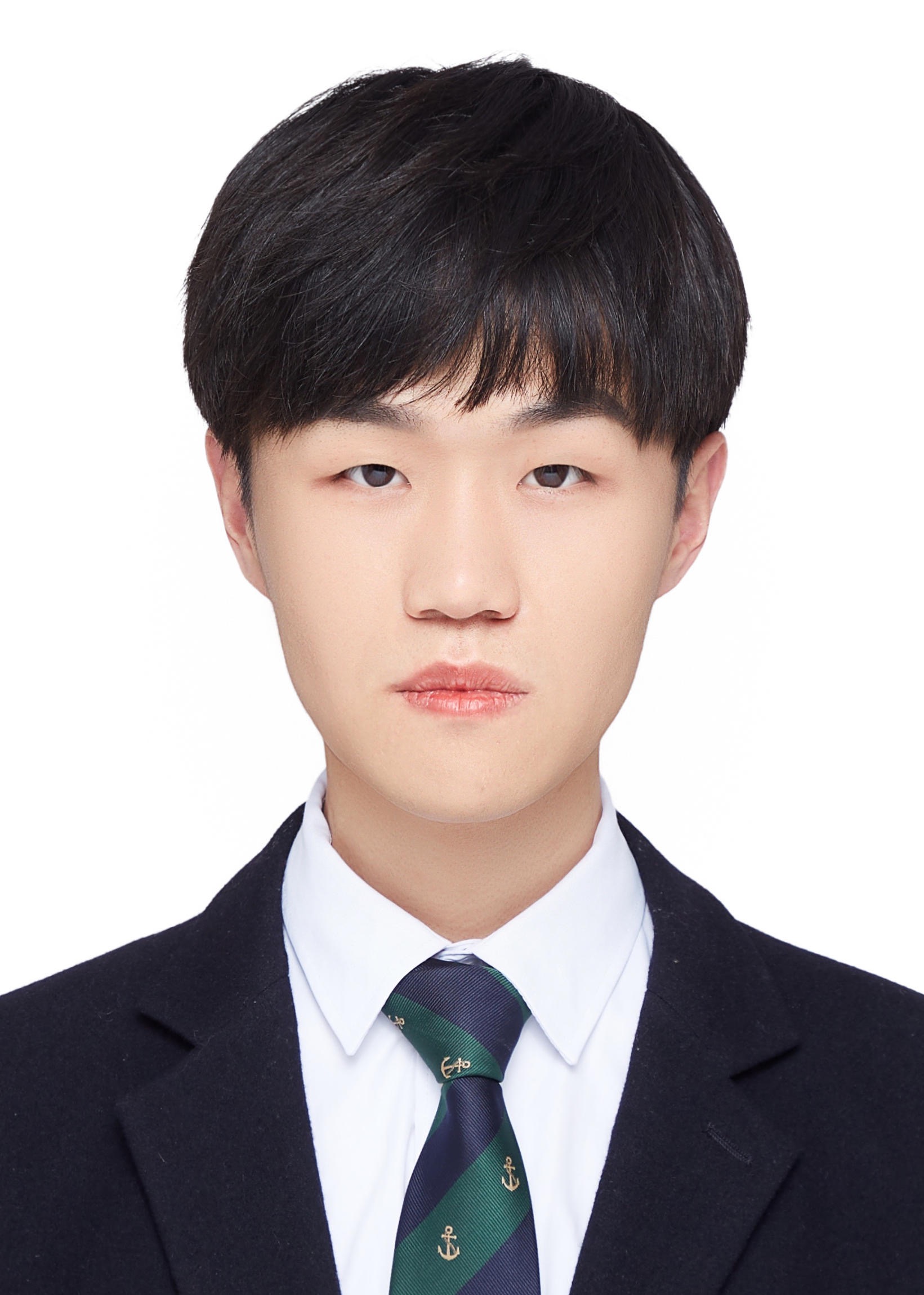}}]{Fangyi Wang}
	(Graduate Student Member, IEEE) received the B.E. degree in communication engineering from Shandong University, Qingdao, China, in 2023. He is currently pursuing the Ph.D.degree in information and communication engineering with the
	Key Laboratory for Information Science of Electromagnetic Waves, School of Information Science and Technology, Fudan University, Shanghai, China.
	
	His research interests include class-incremental learning for HRRP, HRRP image interpretation.
\end{IEEEbiography}
\begin{IEEEbiography}[{\includegraphics[width=1in, height=1.1in,
		trim=0 10pt 0 12pt, clip, keepaspectratio]{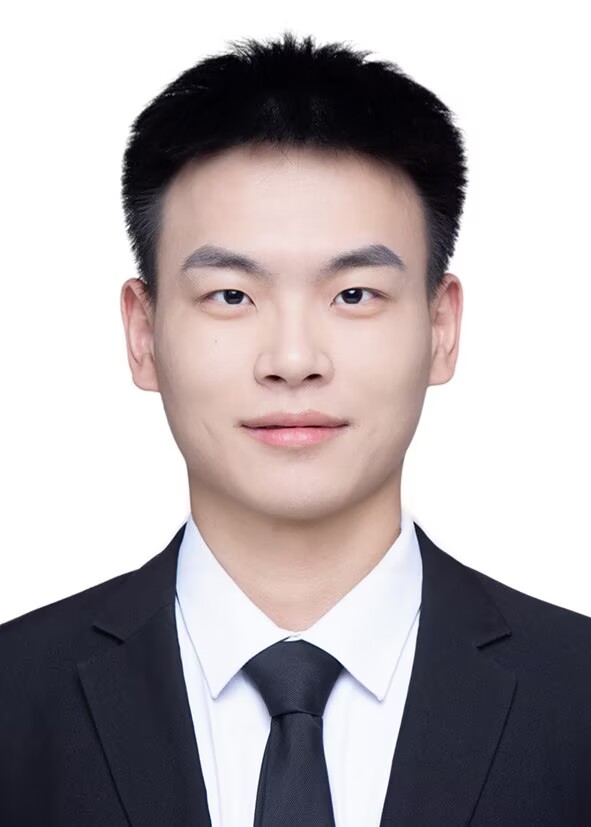}}]{Peng Zhao}
(Graduate Student Member, IEEE) received the B.E. degree in electronic information engineering from Hangzhou Dianzi University, Hangzhou, China, in 2024. He is currently pursuing the M.Eng. degree in electronic information with the Key Laboratory for Information Science of Electromagnetic Waves, School of Information Science and Technology, Fudan University, Shanghai, China. His research interests include intelligent radar signal processingand deep learning for high-resolution range profile (HRRP) interpretation.
\end{IEEEbiography}
\begin{IEEEbiography}[{\includegraphics[width=1in, height=1.1in,
		trim=0 10pt 0 12pt, clip, keepaspectratio]{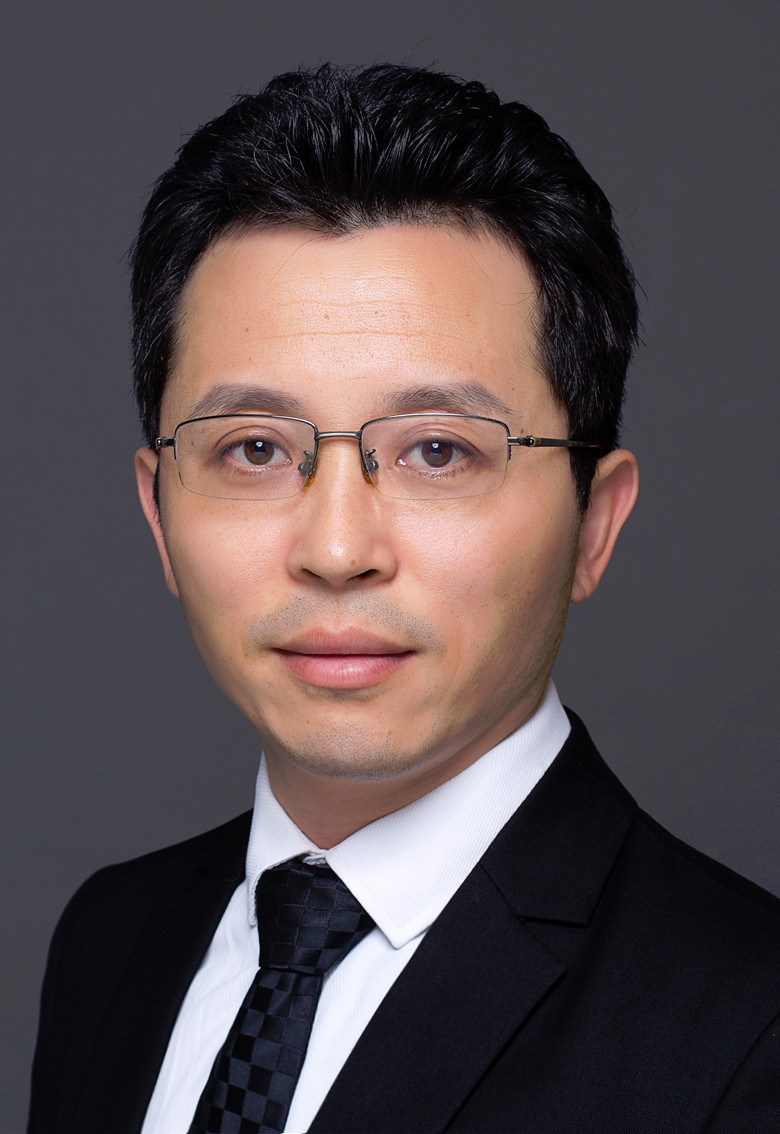}}]{Haipeng Wang}
	(Senior Member, IEEE) received
	the B.S. and M.S. degrees in mechanical and
	electronic engineering from the Harbin Institute of Technology, Harbin, China, in 2001 and 2003, respectively, and the Ph.D. degree in environmental systems engineering from the Kochi University of
	Technology, Kochi, Japan, in 2006.
	
	He was a Visiting Researcher with the Graduate School of Information, Production and Systems, Waseda University, Fukuoka, Japan, in 2008. He is currently a Professor with the Key Laboratory of Electromagnetic Wave Information Science (MoE), Department of Communication Science and Engineering, School of Information Science and Engineering, Fudan University, Shanghai, China. His research interests include signal processing, synthetic aperture radar (SAR) image processing and
	analysis, speckle statistics, and applications to forestry and oceanography,
	and machine learning and its applications to SAR images.
	
	Dr. Wang has been a member of the Technical Program Committee of the
	IEEE Geoscience and Remote Sensing Symposium (IGARSS) since 2011.
	He was a recipient of the Dean Prize of the School of Information Science
	and Engineering, Fudan University, in 2009 and 2017. He also serves as an
	Associate Editor for IEEE GEOSCIENCE AND REMOTE SENSING LETTERS.
\end{IEEEbiography}

\end{document}